%% file: iclr2027_conference.tex
\documentclass{article} %
\PassOptionsToPackage{table}{xcolor} %
\usepackage{iclr2027_conference,times}
\newif\ifarxiv
\arxivtrue                 %
\ifarxiv\iclrfinalcopy\fi

\input{math_commands.tex}

\usepackage{graphicx}
\usepackage{booktabs}
\usepackage{multirow}
\usepackage{tabularx}
\usepackage{xcolor}
\usepackage{amssymb}
\usepackage{subcaption}
\usepackage{placeins}
\graphicspath{{figures/}}

\definecolor{bestred}{RGB}{245,194,193}
\definecolor{secondorange}{RGB}{250,224,195}
\definecolor{thirdyellow}{RGB}{255,251,209}
\newcommand{\vbest}[1]{\cellcolor{bestred}#1}
\newcommand{\vsecond}[1]{\cellcolor{secondorange}#1}
\newcommand{\vthird}[1]{\cellcolor{thirdyellow}#1}
\newcommand{\legendbox}[2]{{\setlength{\fboxsep}{1.5pt}\colorbox{#1}{#2}}}

\newcommand{\ablationheader}{%
  \multirow{2}{*}{\raisebox{-1.2pt}{\textbf{Method}}}
  & \multicolumn{2}{c}{\textbf{Alignment}}
  & \multicolumn{4}{c}{\textbf{Video Quality}}
  & \multicolumn{2}{c}{\textbf{Aesthetic Quality}} \\
  \cmidrule(lr){2-3}\cmidrule(lr){4-7}\cmidrule(lr){8-9}
  & I2VS$\uparrow$ & I2VB$\uparrow$
  & SC$\uparrow$ & BC$\uparrow$ & MS$\uparrow$ & DD$\uparrow$
  & Shadow-v2$\uparrow$ & Waifu-v3$\uparrow$ \\
}
\usepackage{hyperref}
\usepackage{url}
\usepackage[most]{tcolorbox}
\usepackage[T1]{fontenc}  %
\newcommand{\promptfont}{\fontsize{6.5}{7.8}\fontfamily{ComicNeue-TLF}\selectfont}
\definecolor{promptbg}{RGB}{246,246,246}
\definecolor{promptframe}{RGB}{150,150,150}
\tcbset{promptstyle/.style={breakable, listing only, enhanced jigsaw,
  colback=promptbg, colframe=promptframe, boxrule=0.6pt, arc=10pt,
  left=10pt, right=10pt, top=8pt, bottom=8pt, before skip=14pt, after skip=6pt,
  listing options={basicstyle=\promptfont, breaklines=true,
    columns=fullflexible, keepspaces=true, showstringspaces=false,
    breakindent=0pt, literate={-}{{-}}1}}}
\newcommand{\promptfig}[3]{%
  \par\noindent\tcbinputlisting{promptstyle, listing file=prompts/#1.txt}%
  \captionof{figure}{#2}\label{#3}\par\vspace{16pt}}

\definecolor{circred}{RGB}{225,35,35}  %
\DeclareRobustCommand{\circnum}[1]{\tikz[baseline=(c.base)]\node[circle,fill=circred,text=white,inner sep=0pt,minimum size=0.95em,font=\scriptsize\bfseries](c){#1};}

\newcommand{\agenthead}[2]{\raisebox{-0.3em}{\includegraphics[height=1.3em]{#1}}~\textbf{#2}}

\title{Manga2Animate: An Agentic Framework for Faithful Manga Animation with Script and Storyboard Planning}

\author{Sheng-Wei Zhang\thanks{Equal contribution.}\hspace{0.4em}, Hsi-An Chen\footnotemark[1]\hspace{0.4em}, Yan-Ting Liou\footnotemark[1]\hspace{0.4em}, \\
\textbf{Chih-Hsuan Chen, Shih-Chen Tseng, Yu-Lun Liu} \\
Department of Computer Science, National Yang Ming Chiao Tung University\\
}

\begin{document}

\maketitle
\ifarxiv\lhead{Preprint. Under review.}\fi

\begin{figure}[h]
  \centering
  \vspace{-4mm}
  \includegraphics[width=\linewidth]{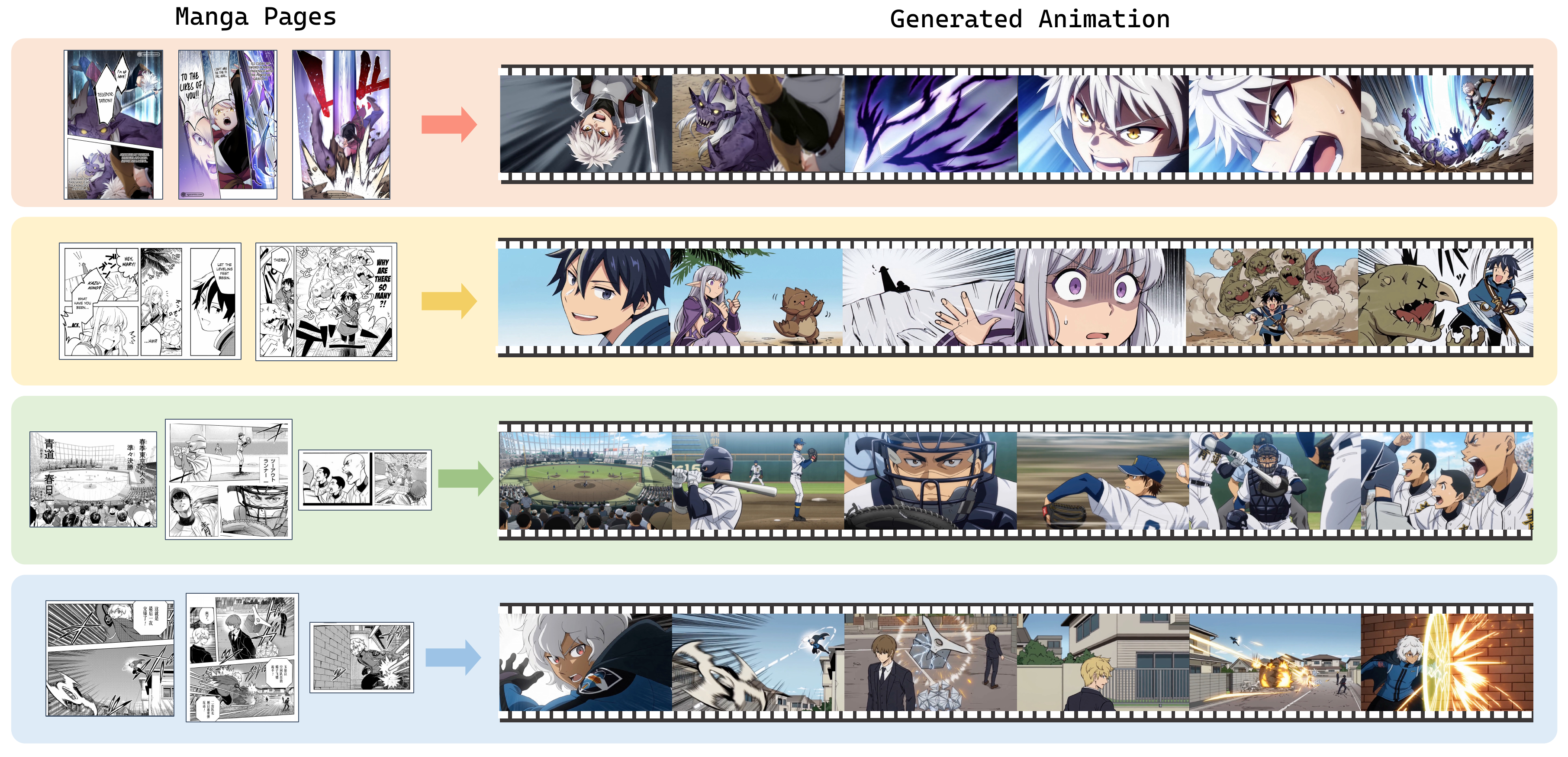}
  \vspace{-4mm}
  \caption{\textbf{Manga2Animate turns existing manga pages into coherent animation that stays faithful to the original manga.} Given a few manga pages \emph{(left)}, our multi-agent framework first builds an animation script and a visual storyboard as intermediate representations, iteratively refining both through creator-reviewer collaboration, and then generates the final animation from them \emph{(right)}. The resulting animation preserves the manga's plot, character identity, and art style, while flowing coherently from clip to clip.}
  \label{fig:teaser}
  \vspace{-4mm}
\end{figure}

\begin{abstract}
\input{sections/0_abstract}
\end{abstract}

\input{sections/1_intro}

\input{sections/2_related}

\input{sections/3_method}

\input{sections/4_experiments}

\input{sections/5_conclusion}

\input{sections/6_statements}
\bibliography{iclr2027_conference}
\bibliographystyle{iclr2027_conference}

\clearpage
\appendix
\setcounter{figure}{0}\renewcommand{\thefigure}{A\arabic{figure}}
\setcounter{table}{0}\renewcommand{\thetable}{A\arabic{table}}
\section*{Appendix}
\input{sections/appendix}

\end{document}

%% file: math_commands.tex
\usepackage{amsmath,amsfonts,bm}

\def\eqref#1{equation~\ref{#1}}

\def\1{\bm{1}}

\DeclareMathAlphabet{\mathsfit}{\encodingdefault}{\sfdefault}{m}{sl}
\SetMathAlphabet{\mathsfit}{bold}{\encodingdefault}{\sfdefault}{bx}{n}

%% file: sections/0_abstract.tex
Adapting existing manga into coherent animation that stays faithful to the source is a valuable yet challenging task. Existing methods either access the manga mainly through text, losing its characters and art style, or feed manga pages directly into video generation models, which misinterpret the manga's structure and produce incorrect storylines. In this paper, we propose \textbf{Manga2Animate}, a multi-agent framework that grounds every stage of generation in the source manga. We decompose the task into two intermediate representations, an animation script and a visual storyboard. Each representation is produced by a creator agent and reviewed against the source panels by a paired reviewer agent. We tailor representation-specific selection and dual-level feedback to progressively refine both before synthesizing the final animation. Extensive experiments and a user study demonstrate that Manga2Animate produces animations more faithful to the source manga and more consistent across shots than state-of-the-art approaches.
Project page: \url{https://annchenn.github.io/Manga2Animate/}

%% file: sections/1_intro.tex
\vspace{-2mm}
\section{Introduction}
\label{sec:intro}
\vspace{-2mm}

\begin{figure}[t]
  \centering
  \includegraphics[width=0.85\linewidth]{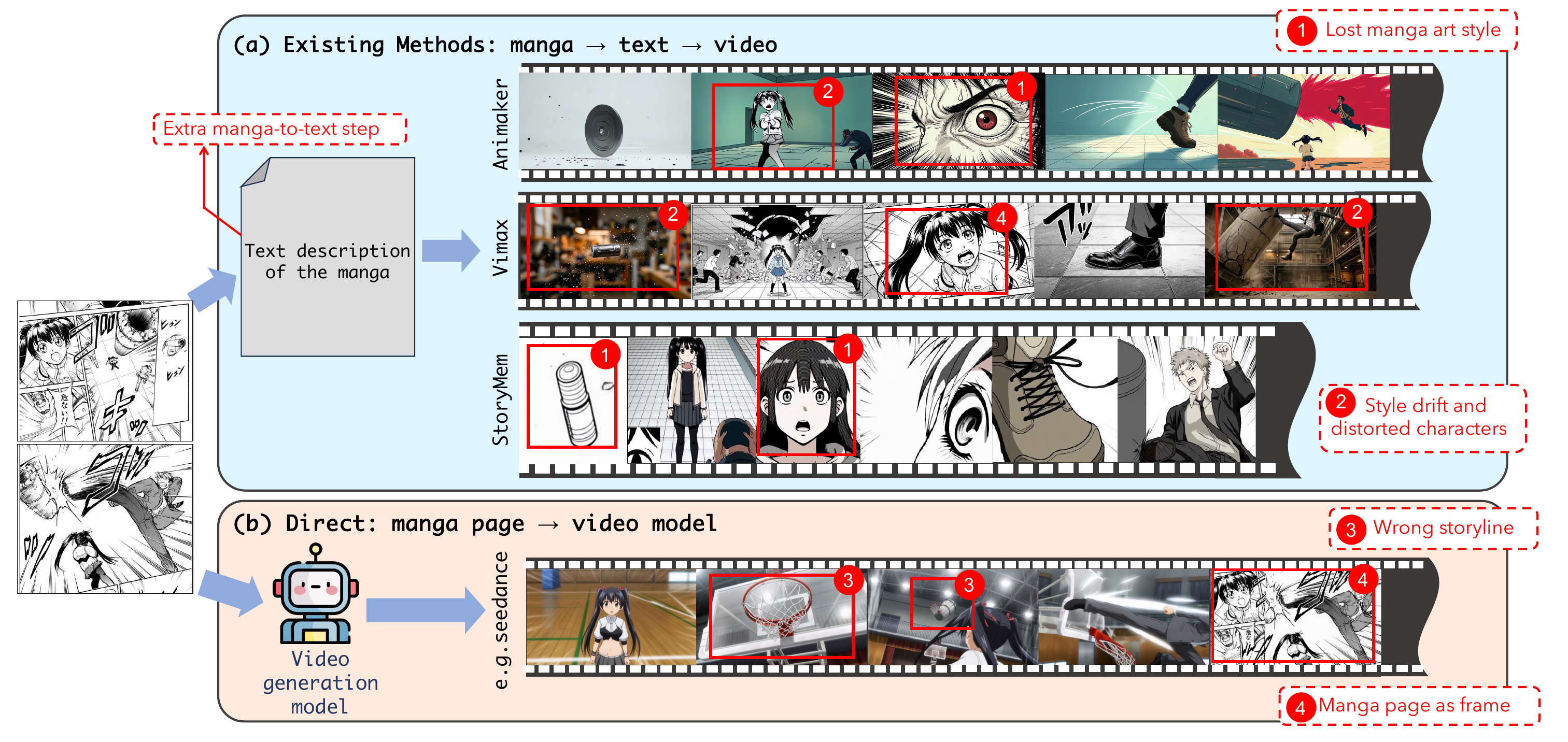}
  \vspace{-2mm}
  \caption{\textbf{Limitations of existing methods to manga-to-animation generation.} (a) Text-based storytelling methods (AniMaker, ViMax, StoryMem) access the manga mainly through text, losing its art style (\circnum{1}) or distorting characters (\circnum{2}). (b) Feeding manga pages into a video model (e.g., Seedance 2.0) tells the wrong story (\circnum{3}) and shows manga panels as video frames (\circnum{4}).}
  \label{fig:limitations}
\end{figure}

Manga is one of the richest sources of stories for animation. Popular titles are routinely adapted into anime, and fans expect these adaptations to move smoothly and tell the story coherently while remaining faithful to the characters and art style they love. Yet producing such an adaptation remains slow and costly, often taking large teams of artists months of storyboarding, key animation, in-betweening, and coloring.

At the same time, a manga already specifies much of what an animation needs, including who the characters are and what they look like, how each shot is framed, and how the plot unfolds. With recent advances in video generation, it is natural to ask whether a manga can be brought to life automatically. However, as Fig.~\ref{fig:limitations} shows, storytelling systems that first describe the manga in words discard its visual appearance, so the original art style is lost (\circnum{1}) and characters change appearance or drift between anime and photorealistic styles (\circnum{2}). Feeding manga pages directly into a video model fails for a different reason. The model knows neither the order in which the panels should be read nor what speech balloons mean, so it tells the wrong story (\circnum{3}) or even places manga panels, or entire pages, directly into the video as frames (\circnum{4}).

Our key observation is that today's video generation models already know how to animate. What they cannot do is understand a manga. Given a manga page, they produce smooth and appealing motion, yet they follow the panels in the wrong order and misread the speech balloons, so the result looks convincing but tells a completely wrong story (Fig.~\ref{fig:limitations}b). We therefore do not replace these models. Instead, we keep their rich visual knowledge and give them what they are missing, a clear understanding of the manga. Much like an animation studio, Manga2Animate (Fig.~\ref{fig:teaser}) first turns the manga into an animation script, which describes what happens in each panel, and a storyboard, which shows how each shot should look. Both are built by AI agents and checked against the original panels, so the video model receives a faithful plan instead of raw manga pages. We find that this plan effectively constrains generation. With the script fixing the plot and the storyboard fixing the characters, composition, and art style, the video model no longer has to interpret the manga on its own and can focus on what it does best, bringing the story to life.

Experiments show that Manga2Animate produces animations that are more faithful to the original characters, art style, and plot than existing storytelling and video generation methods, even when these methods are given the same script as ours. Because our framework guides existing models rather than replacing them, it works with both open-source and commercial image and video generators, and outperforms prior methods with either. In a user study, participants selected our results as the best overall in over 90\% of responses, for both storyboards and videos.

In summary, our contributions are threefold:
\begin{itemize}
  \item We present Manga2Animate, to the best of our knowledge the first automated framework that turns existing manga into animation that is faithful to source manga.
  \item We show that planning through a script and a storyboard, iteratively refined with dual-level feedback from both and verified against the original manga, is key to preserving the characters, art style, and plot.
  \item We establish an evaluation protocol for manga-to-animation generation, under which Manga2Animate outperforms existing methods, and we further validate these results with a user study.
\end{itemize}

%% file: sections/2_related.tex
\vspace{-2mm}
\section{Related Work}
\label{sec:related}

\vspace{-2mm}
\paragraph{Manga understanding.}
Built on annotated corpora~\citep{matsui2017sketch,aizawa2020building,vivoli2024comix,li2024manga109dialog}, manga understanding has progressed from detecting panels, characters, and text~\citep{shen2023maru} to ordering panels and attributing dialogue~\citep{sachdeva2024manga}, naming characters across chapters~\citep{sachdeva2024tails}, and narrating chapters as prose~\citep{sachdeva2025panels}. Vision-language models are now benchmarked on manga~\citep{baek2026mangavqa,ikuta2025mangaub,vivoli2025comicspap}, yet still misread cross-panel narrative~\citep{baranwal2025re} and can rely on memorized popular content rather than visual evidence~\citep{szu2025beyond}; generation from comics remains largely unexplored~\citep{vivoli2024one}. We use~\citet{sachdeva2024manga} for pre-processing and, given this fragility, review every intermediate output against the source panels.

\vspace{-2mm}
\paragraph{Story visualization.}
Story visualization renders a narrative as a consistent image sequence, evolving from GANs~\citep{li2019storygan} to diffusion models with visual memory~\citep{rahman2023make}, identity conditioning~\citep{ye2023ip}, shared attention~\citep{tewel2024training,zhou2024storydiffusion}, frame context~\citep{liu2024intelligent}, multi-subject personalization~\citep{he2025anystory,wang2025ms,wu2025less}, iterative refinement~\citep{mao2026story}, and language-model layout planning~\citep{he2025dreamstory,wang2025autostory,cheng2024theatergen}, with dedicated benchmarks~\citep{zhuang2026vistorybench,gao2025vinabench} and extensions to manga generation~\citep{wu2025diffsensei,wang2026mangaflow}. These methods generate from text, whereas our storyboard is conditioned on and selected against the manga panels.

\vspace{-2mm}
\paragraph{Multi-shot video generation.}
Video diffusion has advanced from short clips~\citep{ho2022video} to foundation models~\citep{kong2024hunyuanvideo,wan2025wan,yang2025cogvideox,polyak2024movie,seedance2026seedance}, whose adherence to explicit specifications is now benchmarked~\citep{huang2026prowbench}. Cross-shot consistency is pursued through long-context tuning~\citep{guo2025long,xiao2026captain,meng2026holocine}, keyframe memory~\citep{zhang2025storymem,zhou2026videomemory}, storyboard anchoring~\citep{zhang2026stage}, and reference-based identity injection~\citep{jiang2025vace,liu2025phantom,hu2025hunyuancustom}. For animation, anime-oriented models and datasets~\citep{jiang2024anisora,qiu2025animeshooter} and in-betweening and colorization methods~\citep{xing2024tooncrafter,meng2025anidoc,huang2024lvcd,li2025tooncomposer,liu2025manganinja,tang2025generative} create motion and color from sparse inputs. We use such models as interchangeable backends and, in the spirit of decoupling explicit state from generative rendering~\citep{huang2026programmable}, obtain consistency from the reviewed script and storyboard rather than from architecture (Sec.~\ref{sec:quantitative}).

\vspace{-2mm}
\paragraph{Agentic storytelling video generation.}
Language and vision-language agents increasingly orchestrate video production from script to storyboard, clips, and assembly~\citep{lin2023videodirectorgpt,li2024anim,xu2025mm,wu2025automated,hu2024storyagent,xu2025filmagent,wang2026mavis,huang2026vimax}, some with critics that select clips by tree search~\citep{shi2025animaker}, gate anime production stages~\citep{zhang2025anime,yan2026animeagent}, or self-improve at test time~\citep{long2026vista}, following self-feedback in language agents~\citep{madaan2023self,shinn2023reflexion}, a pattern also adopted to verify generated images and figures~\citep{huang2026gen,tseng2026one}. Our creator--reviewer design is closest to these systems, but they invent characters, shots, and plot from a prompt and can only check outputs against their own script, whereas our reviewers check against the manga panels. Earlier manga-to-animation systems only moved a virtual camera over panels or added synthesized speech~\citep{cao2016dynamic,gupta2021c2vnet}; to our knowledge, ours is the first to generate faithful multi-shot animation from manga pages.

%% file: sections/3_method.tex
\vspace{-2mm}
\section{Methodology}
\label{sec:method}
\vspace{-2mm}

\begin{figure}[t]
  \centering
  \includegraphics[width=0.88\linewidth]{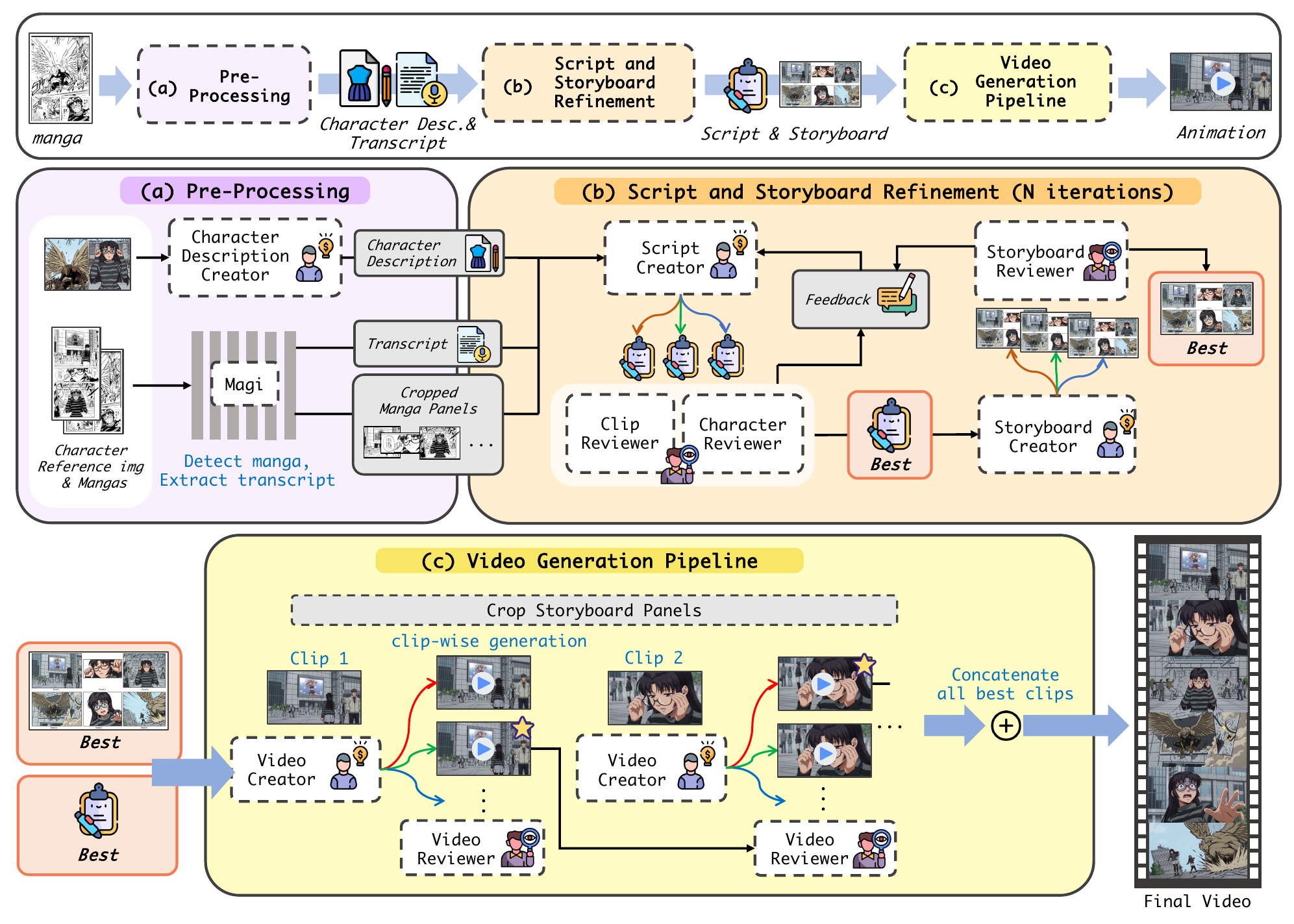}
  \vspace{-2mm}
  \caption{\textbf{Overview of the Manga2Animate pipeline.} (a) Pre-processing extracts ordered panels, a transcript, and character descriptions. (b) Creator and reviewer agents refine an animation script and a visual storyboard over $N$ rounds and select the best script and best storyboard, dual-level feedback also guides the next round. (c) Video generation runs in first-frame or reference-conditioned mode.}
  \label{fig:pipeline}
\end{figure}

\subsection{Task Formulation}
\label{sec:task}
\vspace{-2mm}
Given a set of manga pages $M=\{m_1,\ldots,m_p\}$ in their original reading order, our goal is to generate a coherent animated video $V$. Each manga page consists of multiple \textit{manga panels}, where a panel is an individual visual frame depicting a particular moment, action, or scene. From the input pages, we obtain an ordered sequence of $N_p$ target manga panels $P=\{p_1,\ldots,p_{N_p}\}$ ($N_p=6$ in our experiments) and a transcript $T$. For prominent characters appearing in the selected manga panels, we additionally use character reference images $R$. These references are taken from official color artwork when available; otherwise, clear appearances of the characters are cropped from the input manga.

Our pipeline transforms these inputs through two intermediate representations: an animation script $S=\{s_1,\ldots,s_{N_p}\}$ and a visual storyboard $B=\{b_1,\ldots,b_{N_p}\}$, before producing the final video $V$, as illustrated in Fig.~\ref{fig:pipeline}. We also derive per-character descriptions $D$ from the reference images $R$.

The use of $S$ and $B$ is inspired by real-world animation production workflows, where the narrative is first organized into a script and subsequently translated into a storyboard before animation is produced. Following this workflow, $S$ serves as the textual specification of the animation. Each script clip $s_i$ describes the content corresponding to manga panel $p_i$, including the scene, camera, character positions and actions, expressions, and dialogue. The storyboard $B$ then converts this textual specification into a visual plan, where each storyboard panel $b_i$ specifies the appearance and scene composition of the corresponding animated clip.

Throughout the pipeline, we maintain an explicit correspondence between the source manga and the generated representations as illustrated in Fig.~\ref{fig:correspondence}: manga panel $p_i$, script clip $s_i$, and storyboard panel $b_i$ all describe the same narrative unit. This alignment allows each generated component to be directly traced back to its source manga panel during review and refinement.

\vspace{-2mm}
\subsection{Agents}
\label{sec:agents}
Manga2Animate uses two types of agents at each stage of the pipeline (Fig.~\ref{fig:pipeline}).

\agenthead{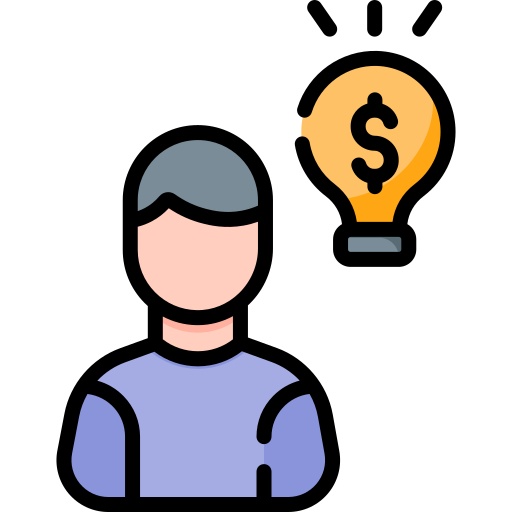}{Creators.} A creator takes the inputs of its stage and produces the intermediate output of that stage.

\agenthead{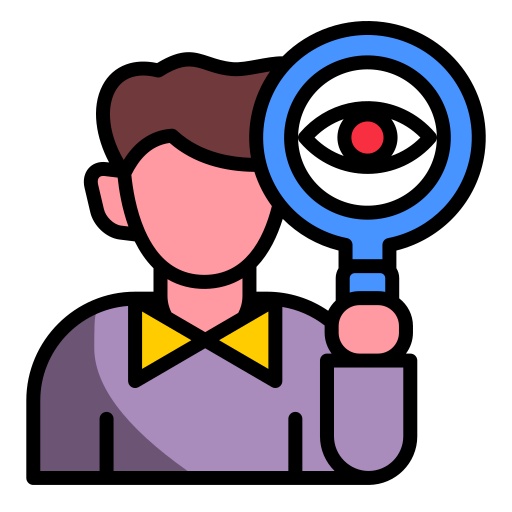}{Reviewers.} A reviewer reviews the outputs of its creator, selects the best one, and gives feedback for refinement. Full review criteria and prompts are listed in Appendices~\ref{app:reviewer} and~\ref{app:prompts}.

\begin{figure}[t]
  \centering
  \includegraphics[width=0.9\linewidth]{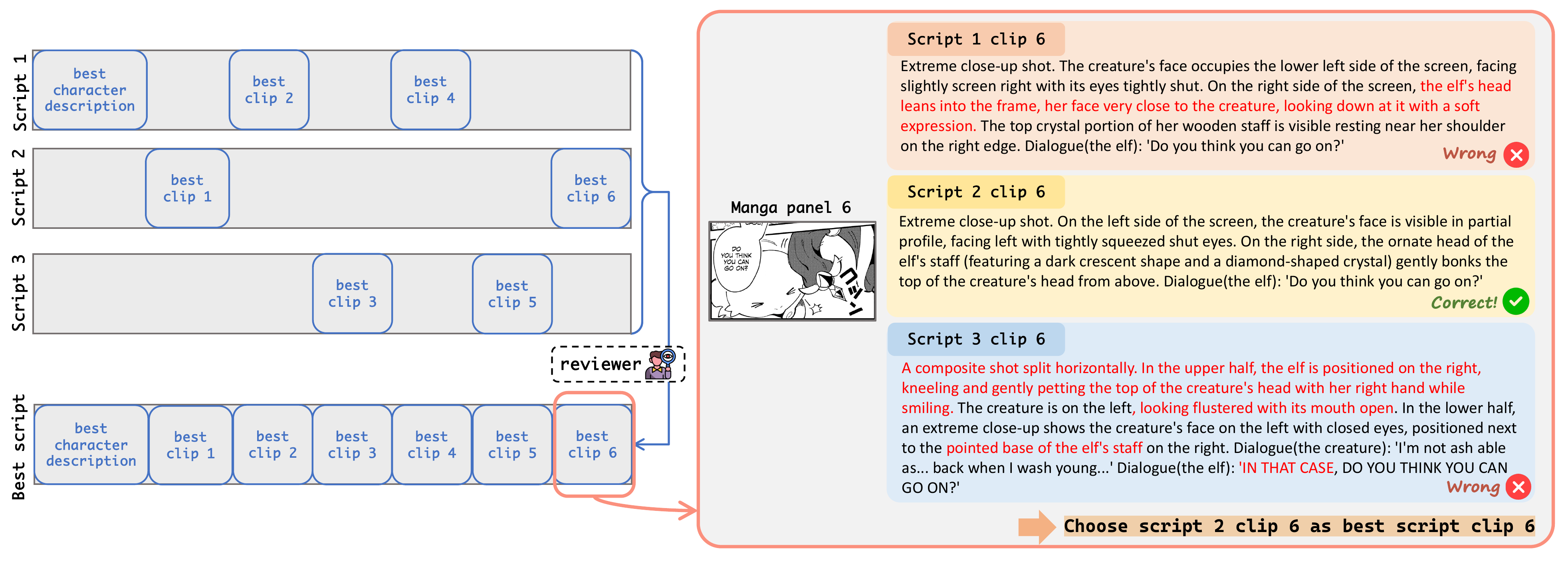}
  \vspace{-2mm}
  \caption{\textbf{Compositional (clip-wise) selection by the Script Reviewer.} The reviewer selects the best character and clip description at each panel position across the $N_s$ candidate scripts and reassembles them into one best script (left). For one panel (right), it flags inaccurate details (red) in two candidates and selects the one matching the panel (green).}
  \label{fig:clipwise}
\end{figure}

\vspace{-2mm}
\subsection{Pre-processing}
\label{sec:preprocess}
Manga pages often exhibit non-trivial panel layouts, making reading order an important prerequisite for preserving the narrative flow. We therefore detect the panels on each page and determine their reading order using a manga-specific detection model and its associated ordering algorithm~\citep{sachdeva2024manga}. The resulting ordered panel sequence is then passed to the subsequent stages of the pipeline. In parallel, the same model also detects and transcribes the text on these pages, which we merge into a single transcript spanning all pages to help the Script Creator write each panel's dialogue and plot (Sec.~\ref{sec:refinement}). Also, we use a Character Description Creator to establish a consistent representation of each character from the reference image $R$. It extracts relatively stable, identity-defining traits, such as facial features and body build, while excluding variable attributes such as clothing and accessories. These character descriptions are then provided to the downstream Script and Storyboard Creators to help maintain character consistency throughout the generation process.

\vspace{-2mm}
\subsection{Script and Storyboard Refinement}
\label{sec:refinement}
To improve generation reliability while controlling API costs, we combine a small number of candidates per stage with iterative refinement, in which reviewer feedback guides the next round of generation. We run $N = 2$ refinement rounds: the first round explores initial candidates, and the second round revises them using the first-round feedback.

In each refinement round, the Script Creator generates $N_s$ script candidates ($N_s = 3$) conditioned on the manga panels, character descriptions, transcript, and, from the second round onward, the feedback from the previous round. Errors in scripts are largely local, a candidate may describe one panel precisely yet misread another, and different candidates tend to make different mistakes. Selecting a single script as a whole would discard the correct descriptions contained in the other candidates. The Script Reviewer therefore performs compositional selection at the clip level (Fig.~\ref{fig:clipwise}), preserving complementary strengths across candidates, and records the remaining weaknesses of each selected component as feedback for the next round.

The Storyboard Creator then generates $N_b$ storyboard candidates ($N_b = 3$) for the selected script. Unlike scripts, closed-source storyboards are selected as a complete image because their panels are generated jointly and share style, color, and lighting; recombining panels across candidates could introduce visual inconsistencies. The Storyboard Reviewer compares candidates pairwise in the first round and, in later rounds, compares each new candidate with the previous best, judging each storyboard against the script that produced it. If no new candidate wins, the previous best storyboard is retained with its script and feedback, keeping the final pair consistent. The open-source backend instead generates and selects panels individually before compositing them due to its constraints (more details in Appendix~\ref{app:opensource}).

Script-level and storyboard-level feedback are jointly provided to the next-round Script Creator, together with the best script from the previous round and the source manga panels. This dual-level feedback allows the pipeline to refine the textual specification using evidence from its visual realization. Issues flagged by the Storyboard Reviewer are treated as symptoms of ambiguous or inaccurate script descriptions: the Script Creator is instructed to identify the responsible part of the script, such as an ambiguous spatial description or a missing detail. To prevent unnecessary changes to already correct content, the Script Creator revises only the script clips or character descriptions explicitly identified as problematic by the reviewers, while leaving all other entries unchanged. When feedback from different reviewers conflicts, the corresponding source manga panel is used as the primary reference for deciding the revision. The script and storyboard selected in the final round are then passed to the video-generation stage.

\vspace{-2mm}
\subsection{Video Generation Pipeline}
\label{sec:videogen}
To accommodate different input formats of video generation models, our video generation pipeline supports two modes: a first-frame mode and a reference-conditioned mode. The former is used for models that take a single image as the first frame, while the latter is used for models that can take the complete storyboard as a reference and generate a multi-shot video at once. Our main experimental results use the reference-conditioned mode.

In first-frame mode, each storyboard panel serves as the first frame of its corresponding clip. For each clip, the Video Creator generates $N_v$ candidates ($N_v = 2$) conditioned on that panel and its script description, and the Video Reviewer selects the better one using the script, panel, and previously selected clip when available. This greedy and clip-by-clip process enforces cross-clip continuity at selection time, after which the selected clips are concatenated in story order. In reference-conditioned mode, the Video Creator instead uses the complete best script and full storyboard to generate $N_v$ complete multi-shot candidates in one pass, from which the Video Reviewer selects the final result.

%% file: sections/4_experiments.tex
\vspace{-2mm}
\section{Experiments}
\label{sec:experiments}

\vspace{-2mm}
\subsection{Datasets}
\label{sec:datasets}
\vspace{-2mm}
Our evaluation set contains eight stories of six ordered manga panels, drawn from Manga109~\citep{matsui2017sketch,aizawa2020building} and additional cases covering diverse backgrounds, characters, interactions, and actions. All methods use the same stories, enabling evaluation of both panel-level fidelity and cross-clip consistency.

\vspace{-2mm}
\subsection{Baselines}
\label{sec:baselines}
\vspace{-2mm}
We use StoryDiffusion~\citep{zhou2024storydiffusion}, Story-Iter~\citep{mao2026story}, and AnyStory~\citep{he2025anystory} as storyboard-generation baselines, and AniMaker~\citep{shi2025animaker}, ViMax~\citep{huang2026vimax}, and StoryMem~\citep{zhang2025storymem} as video-generation baselines. We further compare with direct manga-to-video generation by feeding the manga pages directly into Seedance 2.0 or MiniMax-H3. To compare generation ability rather than manga understanding, every script-based baseline receives the same refined script as Manga2Animate. We also provide the same manga character reference images whenever the baseline accepts reference images. Full adaptation details are in Appendix~\ref{app:baselines}.

\vspace{-2mm}
\subsection{Evaluation Metrics}
\label{sec:metrics}
\vspace{-2mm}
MangaEval adapts existing metrics to the manga input and multi-shot output, evaluating storyboard Visual Consistency, Content Alignment, Aesthetic Quality, and Cost, as well as video Manga Alignment, Video Quality, Aesthetic Quality, and Cost. Full definitions and objectives are in Appendix~\ref{app:metrics} (Tabs.~\ref{tab:metrics_storyboard} and~\ref{tab:metrics_video}).

\vspace{-2mm}
\paragraph{Video Evaluation Metrics.}
At the video level, we evaluate I2V Subject (I2VS), I2V Background (I2VB), Subject Consistency (SC), Background Consistency (BC), Motion Smoothness (MS), and Dynamic Degree (DD) segment by segment against the source panels. We replace VBench++'s~\citep{huang2025vbench++} DINO~\citep{caron2021emerging} with DINOv3~\citep{simeoni2025dinov3} for I2VS and its binary DD with a continuous motion-magnitude score; aesthetics are evaluated on sampled frames with two anime-specific models.

Detailed storyboard evaluation metrics are provided in Appendix~\ref{app:metrics} (Table~\ref{tab:metrics_storyboard}).

\vspace{-2mm}
\subsection{Quantitative Comparisons}
\label{sec:quantitative}
\vspace{-2mm}

\input{tables/video_baseline}

Overall, Table~\ref{tab:video} shows that our pipeline better preserves the manga content and keeps the video clips visually consistent. The improvement remains when we change the image or video model, suggesting that it comes from the script, storyboard, and review process rather than one specific model. DD only measures how much movement a video contains, so a higher DD does not mean that the movement is correct. Storyboard results are reported in Appendix~\ref{app:storyboard_quantitative}.

\vspace{-2mm}
\subsection{Qualitative Analysis}
\label{sec:qualitative}

\begin{figure}[t]
  \centering
  \begin{subfigure}[t]{0.49\linewidth}
    \centering
    \includegraphics[width=\linewidth]{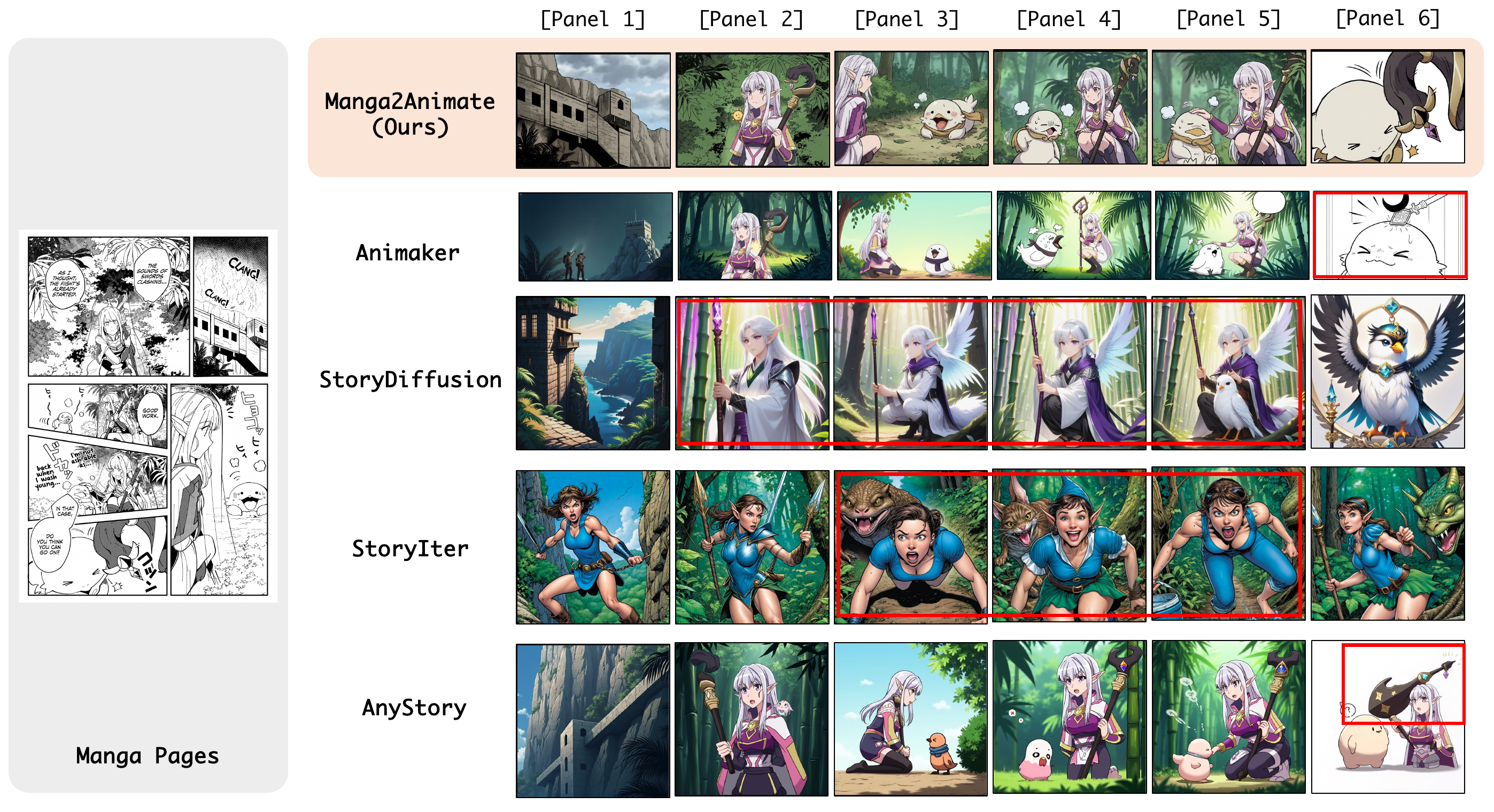}
    \caption{}
    \label{fig:qual_storyboard_a}
  \end{subfigure}\hfill
  \begin{subfigure}[t]{0.49\linewidth}
    \centering
    \includegraphics[width=\linewidth]{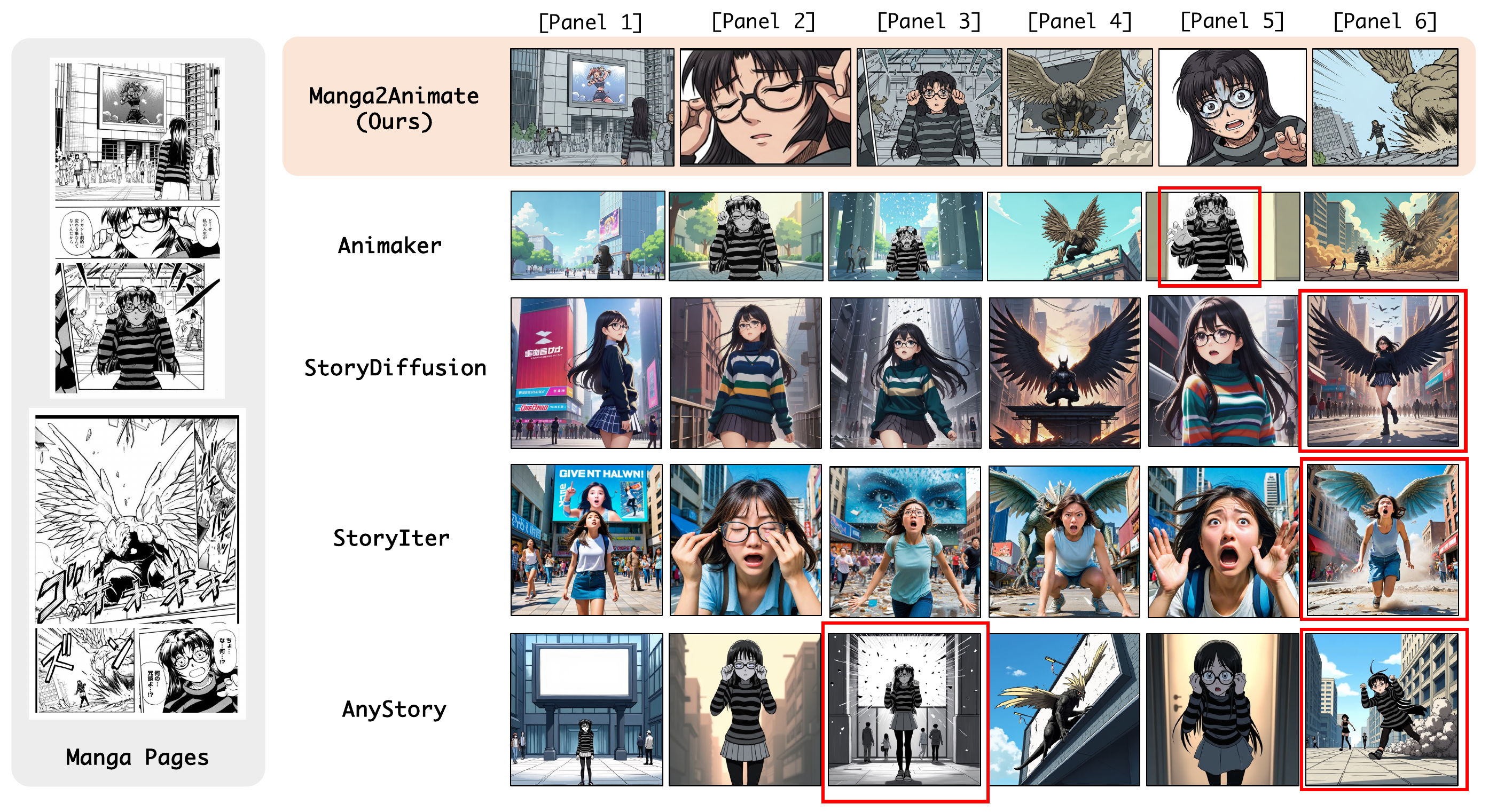}
    \caption{}
    \label{fig:qual_storyboard_b}
  \end{subfigure}
  \vspace{-2mm}
  \caption{\textbf{Qualitative comparison of storyboard generation} on two manga examples (source pages on the left). Manga2Animate keeps the source art style, character identity, and plot across all six panels. AniMaker and AnyStory sometimes leave panels uncolored, StoryDiffusion uses a generic illustration style and repeats content, and Story-Iter becomes photorealistic and changes character appearance. In (b), the baselines also show the falling winged creature incorrectly or leave it out. Red boxes mark example failures.}
  \label{fig:qual_storyboard}
\end{figure}

\vspace{-2mm}
\paragraph{Storyboard Qualitative Results.}
Figure~\ref{fig:qual_storyboard} shows the benefit of generating and reviewing storyboards with the source panels instead of relying only on the shared refined script. Figure~\ref{fig:clipwise} shows how errors are corrected at the script stage.

\begin{figure}[t]
  \centering
  \begin{subfigure}[t]{0.80\linewidth}
    \centering
    \includegraphics[width=\linewidth]{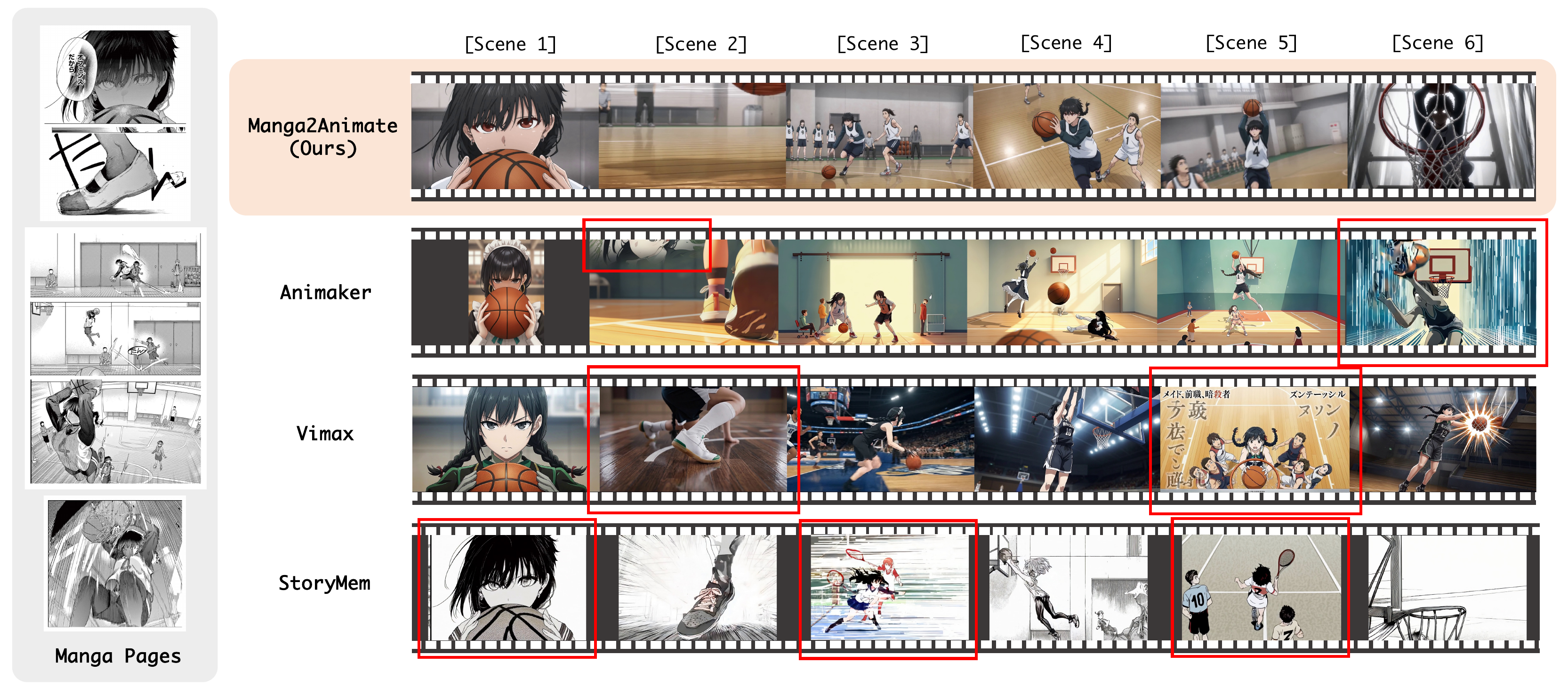}
    \caption{}
    \label{fig:qual_video_a}
  \end{subfigure}\par\smallskip
  \begin{subfigure}[t]{0.80\linewidth}
    \centering
    \includegraphics[width=\linewidth]{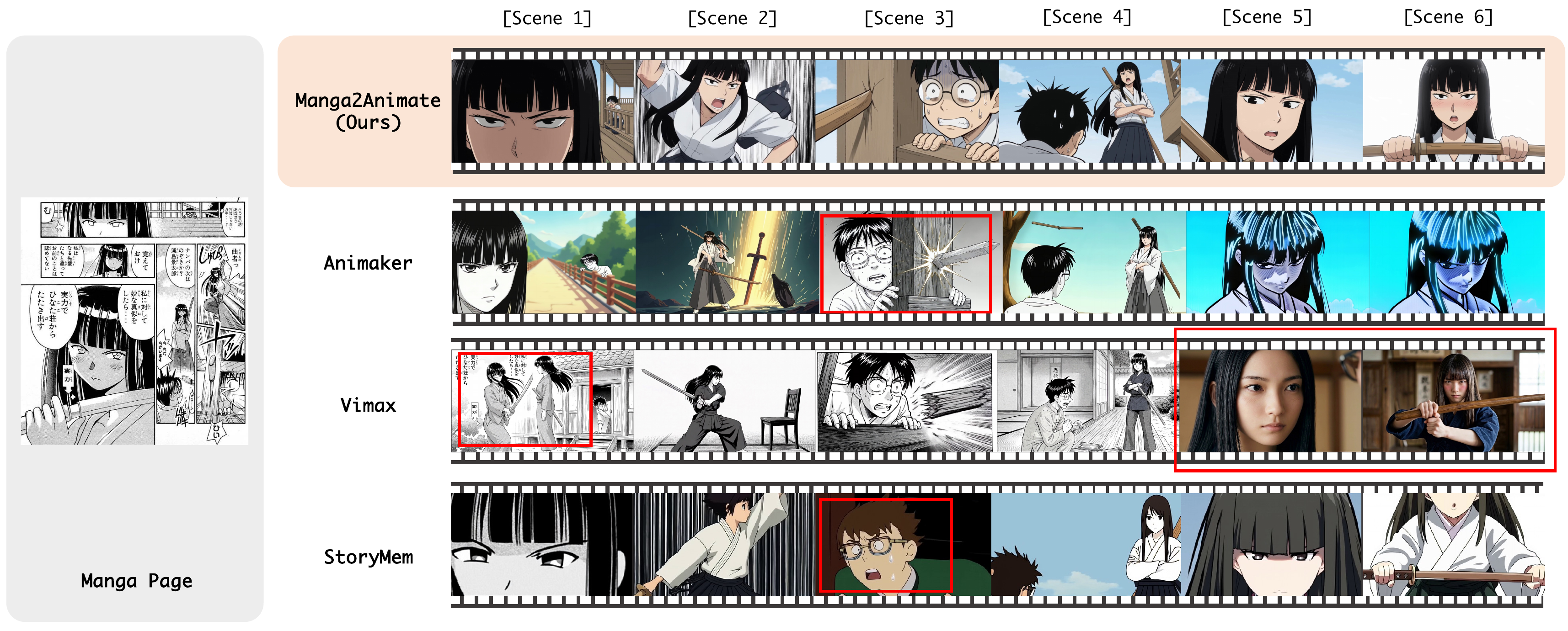}
    \caption{}
    \label{fig:qual_video_b}
  \end{subfigure}
  \vspace{-2mm}
  \caption{\textbf{Qualitative comparison of video generation} on two manga examples (source pages on the left). Manga2Animate keeps the manga's content, characters, and art style. AniMaker adds unrelated people or backgrounds, ViMax mixes anime and photorealistic styles and adds text, and StoryMem produces blurry frames and mistakes the basketball scene for tennis. Baselines may also switch to black-and-white frames or change characters within a video; red boxes mark example failures.}
  \label{fig:qual_video}
\end{figure}

\vspace{-2mm}
\paragraph{Video Qualitative Results.}
As Fig.~\ref{fig:qual_video} shows, the reviewed storyboard provides appearance and composition that match the manga, while the script provides actions and dialogue. Together, they reduce incorrect content even though all methods receive the same refined script. Full videos are in the supplementary material.

\vspace{-2mm}
\subsection{User Study}
\label{sec:userstudy}
\vspace{-2mm}

\begin{figure}[t]
  \centering
  \begin{subfigure}[t]{0.495\linewidth}
    \centering
    \includegraphics[width=\linewidth]{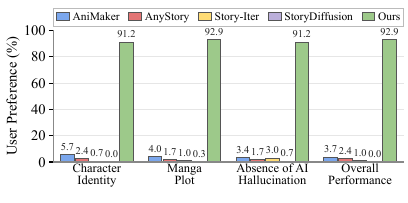}
    \caption{Storyboard}
    \label{fig:userstudy_storyboard}
  \end{subfigure}\hfill
  \begin{subfigure}[t]{0.495\linewidth}
    \centering
    \includegraphics[width=\linewidth]{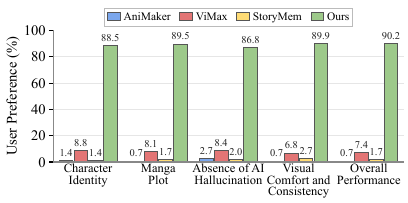}
    \caption{Video}
    \label{fig:userstudy_video}
  \end{subfigure}
  \vspace{-2mm}
  \caption{\textbf{User study on storyboard and video generation.} Thirty-seven participants evaluated eight storyboard and eight video cases shown with the input manga; method names were hidden and outputs randomized. For each criterion, participants selected the best result. Manga2Animate achieves the highest preference on every storyboard criterion (Character Identity, Manga Plot, Absence of AI Hallucination, and Overall Performance) and every video criterion (the same four plus Visual Comfort and Consistency).}
  \label{fig:userstudy}
\end{figure}

As shown in Fig.~\ref{fig:userstudy}, the strong preferences for Manga Plot and Character Identity agree in direction with MangaEval's fidelity metrics (e.g., PA and CIDS), although the human and automatic criteria do not correspond one-to-one.

\vspace{-2mm}
\subsection{Ablation Studies}
\label{sec:ablation}
\vspace{-2mm}
We ablate the three stage-specific reviewers, compare clip-wise with whole-script selection, remove each level of feedback, ablate the refinement loop and storyboard generation, and evaluate video generation mode, using the main configuration as reference.

\input{tables/ablation_reviewer}

\vspace{-2mm}
\paragraph{Stage-Specific Reviewers.}
Table~\ref{tab:abl_reviewer} shows that all three reviewers contribute to the full model's balanced performance.

\vspace{-2mm}
\paragraph{Script Selection Strategy.}
Clip-wise selection improves alignment, consistency, and aesthetics, with only small decreases in I2VB and MS. Full results are in Appendix~\ref{app:additional} (Table~\ref{tab:abl_selection}), and Fig.~\ref{fig:clipwise} shows an example.

\vspace{-2mm}
\paragraph{Dual-Level Feedback.}
Removing either script- or storyboard-level feedback lowers CSD, CIDS, and PA, showing that the two are complementary. Full results are in Appendix~\ref{app:additional} (Table~\ref{tab:abl_feedback}).

\vspace{-2mm}
\paragraph{Refinement Loop and Storyboard Generation.}
Skipping the feedback refinement or the storyboard generation step both lower video quality relative to the full model. Full results are in Appendix~\ref{app:additional} (Table~\ref{tab:abl_pipeline}).

\vspace{-2mm}
\paragraph{Video Generation Mode.}
 These results show that Manga2Animate can use either mode. We use reference-conditioned generation for our main results; full results are in Appendix~\ref{app:additional} (Table~\ref{tab:gemini_seedance_modes}).

%% file: tables/video_baseline.tex
\begin{table}[t]
  \caption{\textbf{Quantitative comparison of multi-shot video generation.} Our main setting (Gemini + Seedance 2.0) performs better than AniMaker, ViMax, and StoryMem in I2VS, I2VB, BC, Shadow-v2, and Waifu-v3. For a fair comparison with ViMax, both methods use the same script, image model, and video model. Our FLUX.2 + MiniMax-H3 setting also performs well, showing that the improvement is not tied to one model combination. Direct generation has higher DD, but it may animate the page layout or produce the wrong actions (Fig.~\ref{fig:limitations}(b)); a higher DD means more movement, not more accurate movement. \legendbox{bestred}{Red}, \legendbox{secondorange}{orange}, and \legendbox{thirdyellow}{yellow} mark the top three results; $\uparrow$/$\downarrow$ show whether higher or lower is better. Call counts are not ranked; -- means no calls.}
  \label{tab:video}
  \vspace{-2mm}
  \centering
  \setlength{\tabcolsep}{2.9pt}
  \resizebox{\linewidth}{!}{%
  \begin{tabular}{llccccccccccc}
    \toprule
    \multirow{2}{*}{\raisebox{-1.2pt}{\textbf{Method}}}
    & \multirow{2}{*}{\raisebox{-1.2pt}{\textbf{Model}}}
    & \multicolumn{2}{c}{\textbf{Alignment}}
    & \multicolumn{4}{c}{\textbf{Video Quality}}
    & \multicolumn{2}{c}{\textbf{Aesthetic Quality}}
    & \multicolumn{3}{c}{\textbf{Cost}} \\
    \cmidrule(lr){3-4}\cmidrule(lr){5-8}\cmidrule(lr){9-10}\cmidrule(lr){11-13}
    & &
    I2VS$\uparrow$ & I2VB$\uparrow$
    & SC$\uparrow$ & BC$\uparrow$
    & MS$\uparrow$ & DD$\uparrow$
    & \mbox{Shadow-v2$\uparrow$} & Waifu-v3$\uparrow$
    & Calls (VLM) & Calls (Video Gen) & Time$\downarrow$ \\
    \midrule
    AniMaker & --
    & 0.775 & 0.749 & 0.922 & 0.950 & \vbest{0.979} & 7.680 & 0.410 & 2.694 & 82.500 & -- & 8h9m \\
    ViMax & Veo 3.1 Fast
    & 0.774 & 0.752 & 0.881 & 0.932 & \vsecond{0.978} & 1.639 & 0.368 & 4.394 & 15.125 & 6.000 & \vthird{12m18s} \\
    ViMax & Seedance 2.0
    & 0.753 & 0.747 & 0.927 & 0.948 & \vthird{0.977} & 2.541 & 0.385 & \vthird{4.764} & 15.125 & 6.000 & 16m33s \\
    ViMax & MiniMax-H3
    & 0.771 & \vthird{0.767} & \vthird{0.938} & \vsecond{0.957} & \vthird{0.977} & 0.977 & 0.326 & 4.505 & 15.500 & -- & 28m58s \\
    StoryMem & --
    & 0.728 & 0.731 & 0.885 & 0.922 & 0.956 & 5.203 & 0.275 & 3.602 & 10.300 & -- & 2h16m8s \\
    \midrule
    Manga $\rightarrow$ Video (Direct) & MiniMax-H3
    & 0.777 & 0.735 & 0.865 & 0.932 & 0.953 & \vsecond{11.787} & 0.402 & 4.611 & -- & -- & \vsecond{5m54s} \\
    Manga $\rightarrow$ Video (Direct) & Seedance 2.0
    & 0.750 & 0.711 & 0.852 & 0.931 & 0.951 & \vbest{13.941} & 0.492 & \vbest{5.250} & -- & 1.000 & \vbest{3m29s} \\
    \midrule
    Ours (FLUX.2) & MiniMax-H3
    & \vbest{0.794} & \vsecond{0.768} & \vbest{0.943} & \vthird{0.953} & 0.968 & 6.872 & \vsecond{0.549} & 4.054 & 59.000 & -- & 2h8m39s \\
    Ours (FLUX.2) & Seedance 2.0
    & 0.781 & 0.741 & 0.892 & 0.942 & 0.958 & 9.377 & \vbest{0.595} & 4.429 & 59.000 & 2.000 & 2h5m46s \\
    Ours (Gemini) & MiniMax-H3
    & \vsecond{0.792} & \vbest{0.791} & \vsecond{0.940} & \vthird{0.953} & 0.969 & 3.715 & 0.413 & 4.352 & 29.500 & -- & 27m5s \\
    Ours (Gemini) & Seedance 2.0
    & \vthird{0.791} & \vsecond{0.768} & 0.931 & \vbest{0.958} & 0.960 & \vthird{9.457} & \vthird{0.518} & \vsecond{5.090} & 29.500 & 2.000 & 24m49s \\
    \bottomrule
  \end{tabular}}
\end{table}

%% file: tables/ablation_reviewer.tex
\begin{table}[t]
  \caption{\textbf{Effect of removing each stage-specific reviewer.} When a reviewer is removed, that stage generates only one candidate and receives no review or feedback; all other settings stay the same. Removing any reviewer lowers the overall alignment and consistency scores. Removing the Script Reviewer causes the largest drop in I2VS. Some variants score higher on MS or Shadow-v2, but score lower on alignment, consistency, or DD. \legendbox{bestred}{Red} and \legendbox{secondorange}{orange} mark the best and second-best results; higher is better.}
  \label{tab:abl_reviewer}
  \vspace{-2mm}
  \centering
  \setlength{\tabcolsep}{4.4pt}
  \resizebox{\linewidth}{!}{%
  \begin{tabular}{lcccccccc}
    \toprule
    \ablationheader
    \midrule
    Without Script Reviewer
    & 0.766 & 0.752 & 0.912 & \vsecond{0.949} & \vsecond{0.963} & \vsecond{8.592} & \vbest{0.523} & 4.789 \\
    Without Storyboard Reviewer
    & \vsecond{0.783} & \vsecond{0.756} & \vsecond{0.923} & 0.948 & \vbest{0.971} & 5.616 & 0.479 & 4.540 \\
    Without Video Reviewer
    & 0.768 & 0.749 & 0.902 & 0.941 & 0.957 & 8.330 & 0.517 & \vsecond{4.996} \\
    Full Model
    & \vbest{0.791} & \vbest{0.768} & \vbest{0.931} & \vbest{0.958} & 0.960 & \vbest{9.457} & \vsecond{0.518} & \vbest{5.090} \\
    \bottomrule
  \end{tabular}}
\end{table}

%% file: sections/5_conclusion.tex
\vspace{-2mm}
\section{Conclusion}
\label{sec:conclusion}
\vspace{-2mm}
We present Manga2Animate, a multi-agent framework for faithfully adapting existing manga into coherent multi-shot animation. We decompose the task into two intermediate representations grounded in the source manga, an animation script and a visual storyboard. Creator and reviewer agents iteratively generate, review, and refine both. Since the two representations fail in different ways, we compose scripts clip by clip, select storyboards as a whole, and feed storyboard-level issues back to the script. Under MangaEval, an evaluation protocol adapted from ViStoryBench and VBench++, Manga2Animate outperforms state-of-the-art storyboard and video generation methods on manga-fidelity and consistency metrics, even when baselines receive the same refined script and character references, and is strongly preferred in a user study. Promising future directions include improving robustness to complex or roughly sketched art styles, developing source-grounded evaluation of story events, and extending narrative understanding across pages and chapters.

\vspace{-2mm}
\paragraph{Limitations.}
Our method has several limitations. First, as an agentic pipeline built on stochastic generative models, its outputs may vary across runs. Second, API-based models may restrict generation of copyrighted or policy-sensitive content. Finally, when the source manga is highly sketchy, cluttered, or compositionally complex, or a scene requires context from elsewhere in the story, the pipeline may misinterpret panel content.

%% file: sections/appendix.tex
This appendix provides additional details that complement the main paper. Sec.~\ref{app:impl} describes the implementation details, including model versions, hyperparameters, video specifications, hardware, and cost. Sec.~\ref{app:opensource} explains the panel-wise generation strategy of the open-source storyboard backend, Sec.~\ref{app:storyboard_quantitative} reports the storyboard quantitative evaluation, Sec.~\ref{app:additional} gives the full script-selection-strategy ablation, the dual-level feedback ablation, an ablation on the refinement loop and storyboard generation, and a comparison of reference-conditioned and first-frame video generation, Sec.~\ref{app:baselines} describes how each baseline is adapted to manga inputs, Sec.~\ref{app:metrics} gives the full definitions of the MangaEval metrics and our modifications, Sec.~\ref{app:reviewer} summarizes the criteria, scoring, and selection rules of every reviewer, and Sec.~\ref{app:prompts} lists the full prompts of all agents. The correspondence among panels, script clips, storyboard panels, and video clips (Sec.~\ref{sec:task}) is illustrated in Fig.~\ref{fig:correspondence}. We further provide an HTML page in the supplementary material showcasing our generated videos.

\section{Implementation Details}
\label{app:impl}
\label{app:correspondence}

\begin{figure}[h]
  \centering
  \includegraphics[width=\linewidth]{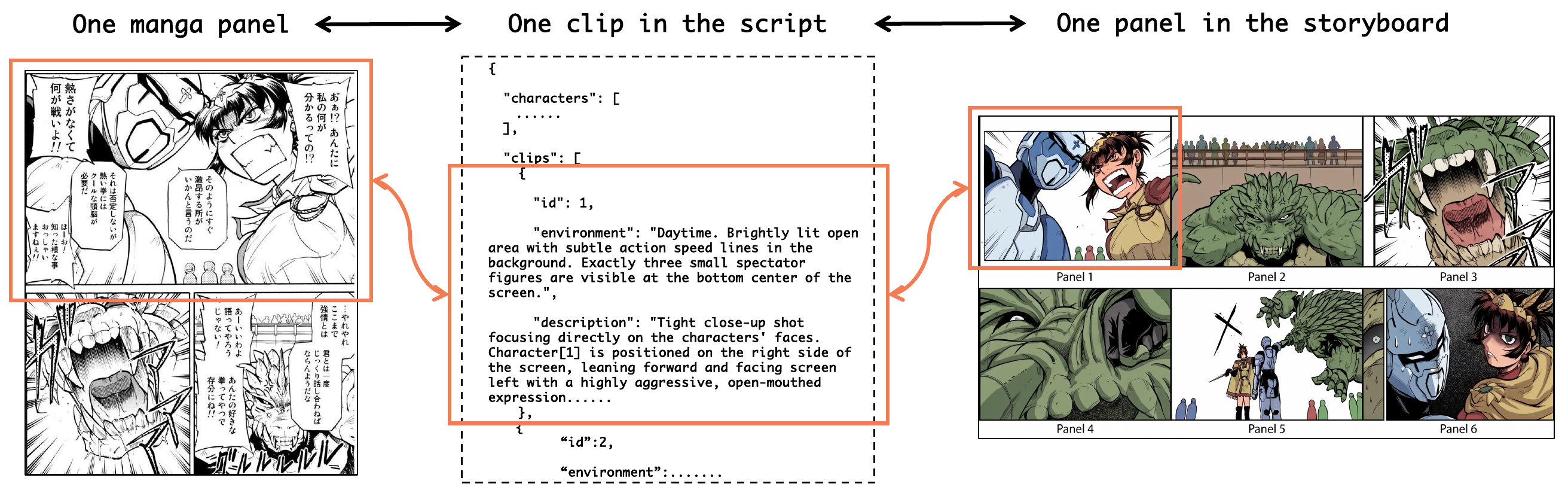}
  \caption{\textbf{Correspondence among intermediate representations.} Each manga panel \emph{(left)} corresponds to one clip in the animation script \emph{(middle)}, which describes the environment, character positions, actions, and expressions of that panel in a structured JSON format. Each script clip is then rendered as one storyboard panel \emph{(right)}. In first-frame mode, that panel conditions the corresponding video clip directly; in reference-conditioned mode, the complete storyboard jointly conditions the multi-shot video.}
  \label{fig:correspondence}
\end{figure}

\paragraph{Models.}
The Character Description Creator, Script Creator, Character Reviewer, Clip Reviewer, Storyboard Reviewer, and open-source Panel Reviewer use gemini-3.7-flash with the thinking level set to HIGH; the Video Reviewer uses gemini-3.1-pro-preview. The closed-source Storyboard Creator uses gemini-3.1-flash-image and outputs a 16:9 storyboard at $1376\times768$ in a $3\times2$ grid. We keep the default sampling parameters, including temperature, for all Gemini calls. Panel detection, reading-order estimation, and OCR use MAGI~\citep{sachdeva2024manga}. The open-source Storyboard Creator uses FLUX.2-dev with 28 inference steps and a guidance scale of 4.0. It generates three candidates per panel at $768\times640$, and its panel reviewer, also gemini-3.7-flash, scores each candidate independently and keeps the highest-scoring one per panel. All experiments use $N = 2$ refinement rounds, $N_s = 3$, $N_b = 3$, and $N_v = 2$.

\paragraph{Video generation.}
In the reference-conditioned mode, the video model receives the complete best script as a text prompt and the full storyboard grid as a single reference image. The closed-source pipeline uses Seedance 2.0 (dreamina-seedance-2-0-260128) in this mode and produces 15\,s videos at $1280\times720$ and 24\,fps, with generated audio. The open-source video-generation configuration supports two video backends. MiniMax-H3 (Ref2VA) runs in the same reference-conditioned mode and produces 15\,s videos at $1344\times768$ and 24\,fps with audio. Index-AniSora V3.2~\citep{jiang2024anisora} runs in the first-frame mode: each cropped storyboard panel is placed as the first frame of its clip, and each clip has 81 frames at 16\,fps and $1280\times720$. Cropped panels are passed to the video model without resizing or padding, and any aspect-ratio adaptation is handled by the video model's own preprocessing. In all cases, two candidate videos are generated and the Video Reviewer selects the better one; when the reviewer judges the two candidates comparable, the first candidate is kept. All main quantitative comparisons in Table~\ref{tab:video} use the reference-conditioned mode; Table~\ref{tab:gemini_seedance_modes} separately compares the reference-conditioned and first-frame modes.

\paragraph{Hardware and cost.}
Gemini and Seedance calls go through hosted APIs. MAGI runs locally on an NVIDIA RTX A6000 (48\,GB). FLUX.2-dev runs on an NVIDIA A100 80GB PCIe, MiniMax-H3 on two NVIDIA RTX PRO 6000 Blackwell Server Edition (96\,GB) GPUs, and Index-AniSora V3.2 on a single NVIDIA RTX PRO 6000 Blackwell Server Edition (96\,GB). For each story, the closed-source pipeline makes 34.5 calls on average in the script and storyboard stages (28.5 Gemini text calls and 6 image-generation calls) and 3 in the video stage (2 Seedance generations and 1 Gemini review), costing about US\$5.65 per story in total, of which US\$4.55 is Seedance. The costs in Table~\ref{tab:storyboard} cover the script and storyboard stages, whereas the VLM calls and time in Table~\ref{tab:video} are counted end-to-end, covering script generation, storyboard generation, video generation, and all reviews, and Calls (Video Gen) counts video-generation calls only. For locally run image and video backends, no external generation API calls are required, although the creator and reviewer agents still use Gemini APIs. Their local computational cost is reflected in the reported wall-clock time. Wall-clock times depend on hardware, parallelism, and API latency and should be regarded as indicative.

\section{Open-Source Storyboard Backend}
\label{app:opensource}
In our experiments, the open-source image model we tested (FLUX.2-dev) did not reliably generate a complete, high-quality multi-panel storyboard in a single pass. We therefore adopt a panel-wise generation strategy, producing multiple candidates for each panel and selecting the best one before composing the final storyboard. For each panel in the selected script, we construct a prompt that incorporates the character descriptions, the corresponding script content, and constraints derived from the original manga panel to preserve its composition, camera viewpoint, and character pose. When character reference crops are available, they are explicitly associated with the corresponding characters during generation.

FLUX.2-dev is conditioned on the source manga panel and the available character references to generate multiple candidates for each panel. A Gemini-based reviewer then scores each candidate independently on manga/script follow (the primary criterion), character identity consistency, art style and colorization quality, and background fidelity, each on a 0--4 scale, and the highest-scoring candidate is retained. The selected panels are then assembled into a $3\times2$ storyboard grid, following the same output format as the closed-source backend to enable evaluation under a unified protocol. Since every panel is conditioned on its source manga panel and the shared character references, these common conditions help maintain cross-panel consistency.

\section{Storyboard Quantitative Evaluation}
\label{app:storyboard_quantitative}

\paragraph{Evaluation metrics.}
We evaluate storyboards with CSD, CIDS, PA, and OCCM from ViStoryBench~\citep{zhuang2026vistorybench}, adapting three of them to manga: CIDS uses SigLIP~\citep{zhai2023sigmoid} instead of CLIP~\citep{radford2021learning} for finer character-crop detail, OCCM's expected character count comes from the source manga panel rather than the script's cast list, and PA is judged by GPT-4.1 instead of Gemini (our pipeline's reviewer) to reduce self-preference bias. Since all script-conditioned methods share the same refined script, PA measures content adherence under an identical specification, while CSD, CIDS, and OCCM (and the Manga Plot criterion of our user study) capture fidelity to the source manga itself; full definitions are in Appendix~\ref{app:metrics}. For Aesthetic Quality, we replace ViStoryBench's general-purpose predictor with two anime-specific models, Aesthetic Shadow v2\footnote{Released by shadowlilac as shadowlilac/aesthetic-shadow-v2, which is no longer available; we use the community mirror at \url{https://huggingface.co/NeoChen1024/aesthetic-shadow-v2-backup}.} and Waifu-v3\footnote{\url{https://huggingface.co/Eugeoter/waifu-scorer-v3}}, trained on anime/illustration imagery and used for training-data curation in production anime models.

\input{tables/storyboard_baseline}

\paragraph{Results.}
Table~\ref{tab:storyboard} compares storyboard generation across methods. The two Manga2Animate configurations jointly achieve the best results on all four manga-fidelity metrics: the open-source FLUX.2 backend reaches the highest CSD and OCCM, while the closed-source Gemini backend achieves the highest CIDS and PA, each improving over the best baseline. Since all baselines use the same refined script as our method, the PA gap reflects content adherence under an identical specification, and this advantage holds for both an open and a closed image backend rather than relying on a specific image model.

In contrast, StoryDiffusion and Story-Iter score higher on the anime-specific aesthetic metrics but considerably lower on manga fidelity (CIDS, PA, OCCM), since both freely generate imagery from text without being constrained by the original characters and composition. This indicates that single-image aesthetics and adaptation fidelity are distinct aspects: an aesthetic predictor assesses the appeal of an individual illustration without considering whether it is consistent with the source. The strengths of Manga2Animate lie in character preservation, content alignment, and cross-panel consistency, which matter most for this task, and participants in our user study also clearly preferred our results overall (Sec.~\ref{sec:userstudy}).

The two backends offer different cost trade-offs: FLUX.2 avoids external image-generation APIs but takes 1\,h 59\,min on local GPUs, while Gemini takes 20\,min 2\,s with 6 image-generation calls. Both are slower than StoryDiffusion and AnyStory but considerably cheaper than AniMaker, another multi-agent pipeline (8\,h 9\,min); users can choose a backend according to their compute resources and API budget.

\section{Additional Ablation Results}
\label{app:additional}

\paragraph{Script selection strategy.} Sec.~\ref{sec:ablation} compares clip-wise and script-wise script selection and reports the headline result; Table~\ref{tab:abl_selection} gives the full metric breakdown. Clip-wise selection outperforms script-wise selection on I2VS, Subject Consistency, Background Consistency, and both aesthetic metrics, raising Waifu-v3 from 4.423 to 5.090 and Shadow-v2 from 0.473 to 0.518, while DD increases from 5.951 to 9.457. I2VB and Motion Smoothness are slightly lower than with script-wise selection (0.768 vs.\ 0.772 and 0.960 vs.\ 0.963), with both gaps within 0.004. We conjecture that clip-wise selection retains the more complete and specific clip descriptions among the candidates, enabling the video model to produce richer motion and visuals, whereas combining descriptions from different candidates may introduce slight variations in scene settings, causing the small drop in I2VB.

\input{tables/ablation_selection}

\paragraph{Dual-level feedback.} To examine the contribution of each feedback source in the refinement loop (Sec.~\ref{sec:refinement}), we withhold either the script-level or the storyboard-level feedback from the next-round Script Creator, while the reviewers still select the best script and storyboard in every round. All variants use the closed-source configuration, i.e., gemini-3.7-flash for all creator and reviewer agents and gemini-3.1-flash-image as the Storyboard Creator, and are evaluated at the storyboard level, where the feedback takes effect directly. Withholding a feedback source changes only the content of the Script Creator's input, not the number of generation or review calls, so every variant has the same computational cost as the full pipeline (Table~\ref{tab:storyboard}); we therefore omit the cost columns. As shown in Table~\ref{tab:abl_feedback}, removing either feedback source lowers CSD (0.605 to 0.597 and 0.588), CIDS (0.808 to 0.786 and 0.770), and PA (3.50 to 3.33 and 3.39) for the variants without script and without storyboard feedback, respectively. For OCCM, removing script feedback lowers the score from 72.2 to 70.8, whereas removing storyboard feedback yields a comparable 72.6. Removing storyboard feedback also lowers both aesthetic metrics more than removing script feedback (Shadow-v2: 0.249 vs.\ 0.334; Waifu-v3: 3.257 vs.\ 3.813), though the without-script-feedback variant slightly exceeds the full model on Waifu-v3 (3.813 vs.\ 3.798). These results indicate that script-level and storyboard-level feedback are complementary, and combining them mainly improves character identity and content alignment.

\input{tables/feedback_ablation}

\paragraph{Refinement loop and storyboard generation.} Skipping the second-round feedback (w/o Refinement) or replacing the generated storyboard with the original manga panels (w/o Storyboard Gen.) both lower I2VS, I2VB, SC, and BC relative to the full model, and removing storyboard generation causes the larger drop, confirming that both stages contribute to the final video quality.

\input{tables/ablation_pipeline}

\paragraph{Video generation mode.}
Manga2Animate supports both reference-conditioned and first-frame video generation. In the former, the video model receives the complete storyboard as a reference image and generates a full video. In the latter, each storyboard panel serves as the first frame of its corresponding clip, and the selected clips are assembled in order.

\input{tables/video_generation_modes}

As shown in Table~\ref{tab:gemini_seedance_modes}, reference-conditioned mode obtains higher I2VS, BC, Shadow-v2, and Waifu-v3 scores, while first-frame mode obtains higher I2VB and slightly higher SC and MS scores. The results demonstrate that the pipeline can also operate with first-frame video generation when full-storyboard reference conditioning is unavailable. We use reference-conditioned mode for the main results.

\section{Baseline Adaptation}
\label{app:baselines}

\paragraph{StoryDiffusion (storyboard).}
StoryDiffusion's official multi-character mechanism takes only plain-text, per-shot prompts that tag which characters appear, with no interface for image-conditioned identity. To keep the comparison fair, we take the same final best script produced by Manga2Animate's own creator-reviewer refinement (Sec.~\ref{sec:refinement}) and reformat character descriptions and per-shot scene/action text into the per-shot prompts its official multi-character mechanism requires. Because StoryDiffusion has no reference-image input, the manga-cropped character images used elsewhere in our evaluation do not apply here; identity relies purely on text description, a limitation of the method when processing image information. Each story yields 6 panels, composed into the same $3\times2$ grid as the other backends for evaluation.

\paragraph{Story-Iter (storyboard).}
Story-Iter generates a sequence of story images from ordered text prompts and maintains cross-image consistency through Global Reference Cross-Attention (GRCA) over previously generated results. To unify the story content, we convert the final refined script of Manga2Animate (Sec.~\ref{sec:refinement}) into six prompts ordered by manga panel, preserving the characters, interactions, scenes, actions, and camera descriptions. Generation follows the official settings with the comic style. Since Story-Iter does not accept external character reference images, we do not add an identity-control module. The six generated images are composed into a $3\times2$ grid in order and scored with MangaEval.

\paragraph{AnyStory (storyboard).}
AnyStory supports personalized story image generation with character reference images. To unify the story content, we convert the final refined script of Manga2Animate (Sec.~\ref{sec:refinement}) into the structured script required by AnyStory, preserving the characters, scenes, actions, interactions, and camera descriptions of the six panels, and we use the same manga-cropped character images as character references. All other stages follow the official AnyStory implementation, without adding the Storyboard Creator or Reviewer of Manga2Animate. The six generated images are composed into a $3\times2$ grid in panel order and scored with MangaEval.

\paragraph{AniMaker (video).}
AniMaker natively takes a plain-text story and self-generates both the script (via an LLM) and the character portrait (via Hunyuan3D~\citep{zhao2025hunyuan3d}). To compare video-generation ability rather than manga-understanding ability, we replace only these two front-end stages: the script is converted from Manga2Animate's own final, refined script (Sec.~\ref{sec:refinement}) into AniMaker's native \texttt{script.json} format, and the character portrait is replaced with the same manga-cropped reference image used by our pipeline. All remaining stages (environment-image generation, scene-prompt synthesis, MCTS-based candidate clip generation and selection, narration, and final assembly) retain AniMaker's original implementation and default hyperparameters, and outputs are scored under the same MangaEval protocol as all baselines.

\paragraph{StoryMem (video).}
StoryMem generates multi-shot videos shot by shot from a structured story script and a visual memory. To unify the story content, we convert the final refined script of Manga2Animate (Sec.~\ref{sec:refinement}) into the StoryMem format, preserving the characters, actions, scenes, and camera descriptions, and adopt the reference-initialized M2V setting, which uses the six manga panels as the initial visual references. Video generation uses the Wan2.2-I2V-A14B~\citep{wan2025wan} backbone with the corresponding StoryMem LoRA checkpoint, and the visual-memory update and shot-by-shot generation follow the official settings. We keep its original mechanism without adding the storyboard, Video Reviewer, or candidate selection of Manga2Animate, and the generated shots are concatenated in script order for evaluation.

\paragraph{ViMax (video).}
ViMax supports a screenplay as input and automatically performs storyboarding, shot planning, character image generation, and video clip generation. To prevent it from rewriting the story from a short concept, we use its Script2Video workflow and convert the final refined script of Manga2Animate (Sec.~\ref{sec:refinement}) into a screenplay input that preserves the characters, scenes, actions, dialogue, camera descriptions, and narrative order. The manga-cropped character images serve as ViMax's character references to maintain appearance consistency. We use gemini-3.6-flash and gemini-3.1-flash-image as the language and image generation models, and Seedance 2.0 (or Veo 3.1 Fast) as the video generation model through external APIs. All other storyboard, shot-planning, and video-generation stages follow the original Script2Video pipeline of ViMax, without adding the storyboard or reviewer modules of Manga2Animate.

\section{Evaluation Details}
\label{app:metrics}

\input{tables/mangaeval_metrics}

We provide full definitions for the metrics summarized in Tables~\ref{tab:metrics_storyboard} and~\ref{tab:metrics_video}.

\paragraph{Storyboard level.}
Visual Consistency is measured by CSD and CIDS, each reporting the average of a cross score (similarity between generated panels and the character reference images) and a self score (similarity among generated panels). CSD~\citep{somepalli2024measuring} uses CSD-CLIP ViT-L style features and is unmodified. CIDS detects characters in the generated panels with GroundingDINO~\citep{liu2024grounding}, assigns them to reference characters by Hungarian matching~\citep{kuhn1955hungarian}, and compares identity features; we replace the original CLIP at 224px with SigLIP (so400m) at 384px, and the cross score is computed over successfully matched character crops. Content Alignment is measured by PA and OCCM. PA uses GPT-4.1 to score each panel on scene, character action, and camera, and each character crop on single-character action, on a 0--4 scale; each item is sampled 5 times at temperature 1.0 and averaged, and PA is the average over the four aspects. The scoring reference is the per-panel text converted from the refined script, which is shared by all script-conditioned methods. $\mathrm{OCCM} = 100 \cdot \exp\!\left(-\frac{|D-E|}{\epsilon + \max(E,1)}\right)$, where $D$ is the number of characters GroundingDINO detects in the generated panel and $E$ is the expected count, now obtained by applying the same detector to the source manga panel rather than taken from the script's cast list. Aesthetic Quality is measured by two anime-specific models: Aesthetic Shadow v2 (a 1.1B-parameter high-/low-quality ViT classifier, score $= P(\text{high quality})$) and Waifu-Scorer v3 (CLIP ViT-L/14 features with an MLP regression head, output in $[0, 10]$), averaged over the six generated panels. Computational Cost is the average number of VLM/LLM calls, image-generation calls, and wall-clock time per story.

\paragraph{Video level.}
Each video is split into six segments corresponding to the six manga panels, and metrics are computed per segment and averaged. Manga Alignment is measured by I2VS and I2VB. I2VS measures the subject similarity between the frames of each segment and the corresponding character reference image using VBench++'s original formula, which combines the maximum similarity to the reference with the mean and minimum similarity between consecutive frames, with its DINO ViT-B/16 backbone replaced by DINOv3 ViT-B/16; segments without characters are excluded. I2VB measures the background similarity between the frames of each segment and the corresponding source manga panel using DreamSim~\citep{fu2023dreamsim} features, with both formula and backbone unmodified. Video Quality is measured by Subject Consistency and Background Consistency (DINO- and CLIP-based frame-to-first-frame and frame-to-previous-frame similarity, respectively, both unmodified VBench++ metrics), Motion Smoothness (an unmodified VBench++ metric based on AMT frame interpolation~\citep{li2023amt}), and Dynamic Degree. We change DD from VBench++'s binary decision based on the top-5\% RAFT~\citep{teed2020raft} flow magnitudes into a continuous motion-magnitude score: each video is resampled to about 8\,fps, RAFT computes optical flow between consecutive frames at the native resolution, and the flow magnitude is averaged over all pixels and frame pairs and normalized by $256/\min(H,W)$. DD measures motion magnitude rather than motion correctness. Aesthetic Quality reuses Aesthetic Shadow v2 and Waifu-Scorer v3 from the storyboard level, applied to every frame per video and averaged. Computational Cost is the end-to-end average number of VLM calls, generation calls, and wall-clock time per story.

\section{Reviewer Criteria}
\label{app:reviewer}
All reviewers are prompted to first describe each candidate with respect to every criterion before reaching a verdict, and to output a structured JSON object containing per-criterion winners or scores, an overall winner, and improvement suggestions for the winner. The suggestions include only criteria on which the winner still shows observable weaknesses and serve as the feedback passed to the next refinement round. The full prompts are listed in Appendix~\ref{app:prompts}.

\paragraph{Character Reviewer.}
Inputs: the ordered manga panels, the character reference image(s), and the character descriptions of all $N_s$ candidate scripts. Criteria (one winner each, one point per win): (1) clothing match, (2) physical appearance accuracy, including facial features, hairstyle, and body proportions, (3) color accuracy with respect to the character reference image, and (4) left/right specificity of asymmetric features such as scars, hair parts, and accessories. The manga panels serve as the ground truth except for colors, which follow the reference image. Ties in total score are broken in the priority order (2) $>$ (1) $>$ (4) $>$ (3).

\paragraph{Clip Reviewer.}
Inputs: one manga panel, the character descriptions (used only to identify characters), and the $N_s$ candidate descriptions of that panel. Criteria (one winner each): (1) small objects associated with each character, penalizing both missing and hallucinated objects, (2) facial expressions and emotions, (3) background accuracy, (4) object placement and orientation, (5) character body orientation and spatial relationships, and (6) dialogue speaker assignment. Ties are broken in the priority order (6) $>$ (5) $>$ (4) $>$ (2) $>$ (1) $>$ (3). The reviewer runs independently for each panel position, and the winning descriptions are reassembled into the final script (Fig.~\ref{fig:clipwise}).

\paragraph{Storyboard Reviewer.}
Inputs: the six manga panels, the character reference image(s), two candidate storyboards A and B, and the script used to generate them. Thirteen criteria, one point each: (1) character identity preservation, (2) overall art-style consistency with the manga, (3) color presence (not primarily black-and-white), (4) consistent panel size in the $3\times2$ layout, (5) only existing characters, (6) narrative order consistency, (7) script alignment, (8) rich background, (9) background consistency across panels, (10) character and object position consistency across panels, (11) panel-internal object layout match with the manga, (12) no dialogue boxes or captions, and (13) left/right consistency of asymmetric character features. Criteria (3), (4), (5), (6), and (12) are judged pass/fail for each storyboard independently, so both storyboards may receive the point; the others are comparative. The storyboard with the higher total wins, and remaining ties are broken by a fixed criterion priority order. From the second round onward, a variant of the prompt compares each new candidate with the previous best storyboard, each judged against the script it was generated from.

\paragraph{Open-source panel reviewer.}
Inputs: the source manga panel, the character references for that panel (possibly none), the corresponding script clip, and one candidate panel. Each candidate is scored independently on a 0--4 scale for (1) manga/script follow, covering action, pose, composition, and framing (the primary criterion), (2) character identity consistency, (3) art style and colorization quality, penalizing black-and-white line art, screentone residue, and speech bubbles, and (4) background and environment fidelity. The candidate with the highest total is kept, with ties broken by criterion (1).

\paragraph{Video Reviewer.}
Inputs: the generation prompt and two candidate videos A and B, plus the previously selected clip for every clip after the first in the first-frame mode. For the first clip, and in the reference-conditioned mode, the criteria and weights are visual realism (0.35), physical commonsense (0.30), and video-prompt alignment (0.35), where alignment also checks left/right directional correctness. For subsequent clips in the first-frame mode, these weights become 0.2 each, and coherency with the previous clip is added with a weight of 0.4. Each criterion is scored 1 (A wins), 0 (B wins), or 0.5 (tie), and a penalty of 0.5 is subtracted for violating hard constraints, such as sudden appearance or disappearance of main objects, unnecessary scene transitions, or unnaturally fast or slow motion. If the weighted scores differ by less than 0.05, the two candidates are judged comparable and the first candidate is kept.

\clearpage
\section{Agent Prompts}
\label{app:prompts}
We list the full prompts of all creator and reviewer agents in pipeline order. Text in braces or angle brackets (e.g., \texttt{\{Target Transcript\}}, \texttt{<script>}) is a placeholder filled in at run time.

\promptfig{create_character_description}{\textbf{Prompt for the Character Description Creator.}}{fig:prompt_chardesc}
\promptfig{create_script_without_feedback}{\textbf{Prompt for the Script Creator (first round, without feedback).}}{fig:prompt_script0}
\promptfig{create_script_with_one_feedback}{\textbf{Prompt for the Script Creator (refinement rounds, with feedback).}}{fig:prompt_script1}
\promptfig{character_reviewer}{\textbf{Prompt for the Character Reviewer.}}{fig:prompt_charrev}
\promptfig{clip_reviewer}{\textbf{Prompt for the Clip Reviewer.}}{fig:prompt_cliprev}
\promptfig{create_storyboard}{\textbf{Prompt for the Storyboard Creator (closed-source backend).}}{fig:prompt_sb}
\promptfig{storyboard_reviewer_same_script}{\textbf{Prompt for the Storyboard Reviewer} (first round, candidates from the same script).}{fig:prompt_sbrev}
\promptfig{storyboard_reviewer_different_script}{\textbf{Prompt for the Storyboard Reviewer} (later rounds, new candidate vs.\ previous best).}{fig:prompt_sbrev2}
\promptfig{panel_candidate_reviewer}{\textbf{Prompt for the open-source panel reviewer.}}{fig:prompt_panelrev}
\promptfig{video_generation_storyboard}{\textbf{Prompt for the Video Creator (reference-conditioned mode).}}{fig:prompt_video_ref}
\promptfig{video_generation}{\textbf{Prompt for the Video Creator (first-frame mode).}}{fig:prompt_video_kf}
\promptfig{video_eval_first}{\textbf{Prompt for the Video Reviewer} (reference-conditioned mode and the first clip in the first-frame mode).}{fig:prompt_videorev}
\promptfig{video_eval_sequential}{\textbf{Prompt for the Video Reviewer} (subsequent clips in the first-frame mode).}{fig:prompt_videorev_seq}

%% file: tables/storyboard_baseline.tex
\begin{table}[t]
  \caption{\textbf{Quantitative comparison of storyboard generation.} \legendbox{bestred}{Red}, \legendbox{secondorange}{orange}, and \legendbox{thirdyellow}{yellow} cells denote the best, second-best, and third-best results; $\uparrow$/$\downarrow$ indicate that higher/lower is better. Call counts are not ranked; -- denotes no calls of that type.}
  \label{tab:storyboard}
  \centering
  \setlength{\tabcolsep}{3.2pt}
  \resizebox{\linewidth}{!}{%
  \begin{tabular}{lccccccccc}
    \toprule
    \multirow{2}{*}{\raisebox{-1.2pt}{\textbf{Method}}}
    & \multicolumn{2}{c}{\textbf{V.C.}}
    & \multicolumn{2}{c}{\textbf{C.A.}}
    & \multicolumn{2}{c}{\textbf{A.Q.}}
    & \multicolumn{3}{c}{\textbf{Cost}} \\
    \cmidrule(lr){2-3}\cmidrule(lr){4-5}\cmidrule(lr){6-7}\cmidrule(lr){8-10}
    & CSD$\uparrow$ & CIDS$\uparrow$
    & PA$\uparrow$ & OCCM$\uparrow$
    & \mbox{Shadow-v2$\uparrow$} & Waifu-v3$\uparrow$
    & Calls (VLM) & Calls (Img Gen) & Time$\downarrow$ \\
    \midrule
    StoryDiffusion
    & \vthird{0.509} & 0.713 & 2.05 & 57.5 & \vbest{0.697} & \vsecond{5.789} & -- & -- & \vbest{49s} \\
    Story-Iter
    & 0.486 & 0.652 & 1.85 & 66.3 & \vsecond{0.615} & \vbest{6.308} & 60.000 & 11.000 & \vthird{13m20s} \\
    AnyStory
    & \vsecond{0.605} & \vthird{0.792} & 2.66 & \vthird{69.3} & \vthird{0.606} & 3.781 & 10.000 & 1.000 & \vsecond{2m7s} \\
    AniMaker
    & 0.508 & \vsecond{0.797} & \vsecond{3.34} & 67.5 & 0.448 & \vthird{4.522} & 82.500 & 18.000 & 8h9m \\
    \midrule
    Ours (FLUX.2)
    & \vbest{0.622} & \vsecond{0.797} & \vthird{3.14} & \vbest{73.5} & 0.552 & 3.423 & 58.000 & -- & 1h59m \\
    Ours (Gemini 3.1 Image)
    & \vsecond{0.605} & \vbest{0.808} & \vbest{3.50} & \vsecond{72.2} & 0.350 & 3.798 & 28.500 & 6.000 & 20m2s \\
    \bottomrule
  \end{tabular}}
\end{table}

%% file: tables/ablation_selection.tex
\begin{table}[t]
  \caption{\textbf{Ablation on script selection strategy.} We compare Manga2Animate with clip-wise and script-wise selection. The full model selects the best clip description for each manga panel in a clip-wise manner and composes them into the final script, whereas script-wise selection directly picks one complete candidate script. All other models and settings are unchanged. \legendbox{bestred}{Red} cells denote the better result in each column.}
  \label{tab:abl_selection}
  \centering
  \setlength{\tabcolsep}{4.4pt}
  \resizebox{\linewidth}{!}{%
  \begin{tabular}{lcccccccc}
    \toprule
    \ablationheader
    \midrule
    Script-wise Selection
    & 0.785 & \vbest{0.772} & 0.928 & 0.956 & \vbest{0.963} & 5.951 & 0.473 & 4.423 \\
    Clip-wise Selection (Full Model)
    & \vbest{0.791} & 0.768 & \vbest{0.931} & \vbest{0.958} & 0.960 & \vbest{9.457} & \vbest{0.518} & \vbest{5.090} \\
    \bottomrule
  \end{tabular}}
\end{table}

%% file: tables/feedback_ablation.tex
\begin{table}[t]
  \caption{\textbf{Ablation on dual-level feedback.} The reviewers still select the best script and storyboard, but the feedback of one reviewer is withheld from the next-round Script Creator. All variants use the closed-source (Gemini) configuration. The Full Model row is the run reported in Table~\ref{tab:storyboard}. Cost is identical to the full pipeline and is therefore omitted. \legendbox{bestred}{Red} and \legendbox{secondorange}{orange} mark the best and second-best results; higher is better.}
  \label{tab:abl_feedback}
  \centering
  \setlength{\tabcolsep}{4.4pt}
  \small
  \begin{tabular}{lcccccc}
    \toprule
    \multirow{2}{*}{\raisebox{-1.2pt}{\textbf{Method}}}
    & \multicolumn{2}{c}{\textbf{V.C.}}
    & \multicolumn{2}{c}{\textbf{C.A.}}
    & \multicolumn{2}{c}{\textbf{A.Q.}} \\
    \cmidrule(lr){2-3}\cmidrule(lr){4-5}\cmidrule(lr){6-7}
    & CSD$\uparrow$ & CIDS$\uparrow$
    & PA$\uparrow$ & OCCM$\uparrow$
    & \mbox{Shadow-v2$\uparrow$} & Waifu-v3$\uparrow$ \\
    \midrule
    Without Script Feedback
    & \vsecond{0.597} & \vsecond{0.786} & 3.33 & 70.8 & \vsecond{0.334} & \vbest{3.813} \\
    Without Storyboard Feedback
    & 0.588 & 0.770 & \vsecond{3.39} & \vbest{72.6} & 0.249 & 3.257 \\
    Full Model
    & \vbest{0.605} & \vbest{0.808} & \vbest{3.50} & \vsecond{72.2} & \vbest{0.350} & \vsecond{3.798} \\
    \bottomrule
  \end{tabular}
\end{table}

%% file: tables/ablation_pipeline.tex
\begin{table}[ht]
  \caption{\textbf{Ablation on the refinement loop and storyboard generation.} \emph{w/o Refinement} uses the first-pass script and storyboard directly, skipping the second-round feedback; \emph{w/o Storyboard Gen.} keeps the same fully refined script but replaces the generated storyboard with a composite of the original manga panels. Removing either stage lowers I2VS, I2VB, SC, and BC relative to the full model, and removing storyboard generation causes the larger drop. Some variants score higher on MS or Shadow-v2, but lower on alignment, consistency, or DD. \legendbox{bestred}{Red} and \legendbox{secondorange}{orange} mark the best and second-best results; higher is better.}
  \label{tab:abl_pipeline}
  \centering
  \setlength{\tabcolsep}{4.4pt}
  \resizebox{\linewidth}{!}{%
  \begin{tabular}{lcccccccc}
    \toprule
    \ablationheader
    \midrule
    w/o Refinement
    & \vsecond{0.778} & \vsecond{0.754} & \vsecond{0.914} & \vsecond{0.952} & \vbest{0.965} & 7.872 & \vbest{0.521} & 4.818 \\
    w/o Storyboard Gen.
    & 0.766 & 0.745 & 0.896 & 0.947 & \vsecond{0.961} & \vsecond{8.289} & 0.515 & \vsecond{5.050} \\
    Full Model
    & \vbest{0.791} & \vbest{0.768} & \vbest{0.931} & \vbest{0.958} & 0.960 & \vbest{9.457} & \vsecond{0.518} & \vbest{5.090} \\
    \bottomrule
  \end{tabular}}
\end{table}

%% file: tables/video_generation_modes.tex
\begin{table}[h]
  \centering
  \small
  \setlength{\tabcolsep}{4pt}
  \caption{\textbf{Comparison of Video Generation Modes.}
  Both modes use Gemini for storyboard generation and Seedance 2.0 for video generation. Bold indicates the higher value for each metric. Reference-conditioned generation gives better subject alignment, background consistency, and aesthetic scores, while first-frame generation gives better background alignment and slightly better subject consistency and motion smoothness.}
  \label{tab:gemini_seedance_modes}

  \begin{tabular}{l*{8}{c}}
    \toprule
    \multirow{2}{*}{\raisebox{-1.2pt}{\textbf{Mode}}}
      & \multicolumn{2}{c}{\textbf{Alignment}}
      & \multicolumn{4}{c}{\textbf{Video Quality}}
      & \multicolumn{2}{c}{\textbf{Aesthetic Quality}} \\
    \cmidrule(lr){2-3}\cmidrule(lr){4-7}\cmidrule(lr){8-9}
      & I2VS$\uparrow$ & I2VB$\uparrow$
      & SC$\uparrow$ & BC$\uparrow$ & MS$\uparrow$ & DD$\uparrow$
      & \mbox{Shadow-v2$\uparrow$} & Waifu-v3$\uparrow$ \\
    \midrule
    Reference-conditioned
      & \textbf{0.791} & 0.768
      & 0.931 & \textbf{0.958} & 0.960 & \textbf{9.457}
      & \textbf{0.518} & \textbf{5.090} \\
    First-frame
      & 0.772 & \textbf{0.778}
      & \textbf{0.935} & 0.941 & \textbf{0.962} & 4.045
      & 0.302 & 4.042 \\
    \bottomrule
  \end{tabular}
\end{table}

%% file: tables/mangaeval_metrics.tex
\begin{table}[h]
  \scriptsize
  \setlength{\tabcolsep}{2.5pt}
  \begin{minipage}[t]{0.495\linewidth}
    \centering
    \caption{\textbf{Storyboard Evaluation Metrics.}}
    \label{tab:metrics_storyboard}
    \begin{tabularx}{\linewidth}{>{\raggedright\arraybackslash}p{1.45cm}>{\raggedright\arraybackslash}p{1.95cm}>{\raggedright\arraybackslash}X}
      \toprule
      \textbf{Domain} & \textbf{Metric} & \textbf{Brief Description} \\
      \midrule
      \textbf{Visual Consistency}\newline\textbf{(V.C.)}
        & CLIP Style Disentanglement\newline(CSD)
        & Average source-to-generated and inter-panel style consistency \\ \addlinespace[3pt]
        & Character Identification Similarity\newline(CIDS)
        & Average source-to-generated and inter-panel character identity consistency using SigLIP features \\
      \midrule
      \textbf{Content Alignment}\newline\textbf{(C.A.)}
        & Prompt Alignment\newline(PA)
        & GPT-4.1 judgment of scene, shot, character interaction, and individual action adherence \\ \addlinespace[3pt]
        & Onstage Character Count Matching\newline(OCCM)
        & Agreement between detected character counts in generated and source manga panels \\
      \midrule
      \textbf{Aesthetic Quality}\newline\textbf{(A.Q.)}
        & Aesthetic Shadow V2\newline(Shadow-v2)
        & General aesthetic quality of the generated storyboard \\ \addlinespace[3pt]
        & Waifu Scorer V3\newline(Waifu-v3)
        & Anime-oriented visual quality of the generated storyboard \\
      \midrule
      \textbf{Cost}
        & Calls (VLM)
        & Number of VLM calls required for generation and review \\ \addlinespace[3pt]
        & Calls (Img Gen)
        & Number of external image-generation API calls \\ \addlinespace[3pt]
        & Time
        & Total storyboard generation time \\
      \bottomrule
    \end{tabularx}
  \end{minipage}\hfill
  \begin{minipage}[t]{0.495\linewidth}
    \centering
    \caption{\textbf{Video Evaluation Metrics.}}
    \label{tab:metrics_video}
    \begin{tabularx}{\linewidth}{>{\raggedright\arraybackslash}p{1.45cm}>{\raggedright\arraybackslash}p{1.95cm}>{\raggedright\arraybackslash}X}
      \toprule
      \textbf{Domain} & \textbf{Metric} & \textbf{Brief Description} \\
      \midrule
      \textbf{Alignment}
        & I2V Subject Similarity\newline(I2VS)
        & Subject similarity between the source manga and generated video frames using DINOv3 features \\ \addlinespace[3pt]
        & I2V Background Similarity\newline(I2VB)
        & Background similarity between the source manga and generated video frames \\
      \midrule
      \textbf{Video Quality}\newline\textbf{(V.Q.)}
        & Subject Consistency\newline(SC)
        & Temporal consistency of subjects across generated video frames \\ \addlinespace[3pt]
        & Background Consistency\newline(BC)
        & Temporal consistency of backgrounds across generated video frames \\ \addlinespace[3pt]
        & Motion Smoothness\newline(MS)
        & Temporal smoothness of motion across adjacent video frames \\ \addlinespace[3pt]
        & Dynamic Degree\newline(DD)
        & Continuous optical-flow-based motion intensity, normalized to a 256-pixel short side \\
      \midrule
      \textbf{Aesthetic Quality}\newline\textbf{(A.Q.)}
        & Aesthetic Shadow V2\newline(Shadow-v2)
        & General aesthetic quality of the generated video \\ \addlinespace[3pt]
        & Waifu Scorer V3\newline(Waifu-v3)
        & Anime-oriented visual quality of the generated video \\
      \midrule
      \textbf{Cost}
        & Calls (VLM)
        & Number of VLM calls required for generation and review \\ \addlinespace[3pt]
        & Calls (Video Gen)
        & Number of external video-generation API calls \\ \addlinespace[3pt]
        & Time
        & Total video generation time \\
      \bottomrule
    \end{tabularx}
  \end{minipage}
\end{table}